\documentclass[journal]{IEEEtran}
\ifCLASSINFOpdf
\else
\fi
\usepackage{amsmath}
\usepackage{amssymb}
\usepackage{cite}
\usepackage{color}
\usepackage{booktabs}
\usepackage{graphicx}
\usepackage{subfigure}
\usepackage{multirow}
\usepackage{color}

\begin{document}
%
\title{End-to-End Comparative Attention Networks for Person Re-identification}
%
%
%

\author{Hao~Liu, Jiashi Feng, Meibin Qi, Jianguo Jiang
        and Shuicheng Yan, \IEEEmembership{Fellow,~IEEE}
\thanks{H. Liu, M. Qi and J. Jiang are with School of Computer and Information, Hefei University of Technology, Hefei, Anhui 230009 China (e-mail: hfut.haoliu@gmail.com; qimeibin@163.com; jgjiang@hfut.edu.cn).}
\thanks{J. Feng is with the Department of Electrical and Computer Engineering, National University of Singapore, Singapore 119077 (e-mail: elefjia@nus.edu.sg).}%
\thanks{S. Yan is with the Artificial Intelligence Institute, Qihoo 360 Technology Company, Ltd., Beijing 100015, China (e-mail: eleyans@nus.edu.sg).}
\thanks{Manuscript received June, 2016; revised 2016. This work was supported in part by the National Natural Science Foundation of China (Grant Nos. 61371155, 61174170, 61632007.)  and China Scholarship Council (Grant No. 201506690007).}}

\maketitle

\begin{abstract}
Person re-identification across disjoint camera views has been widely applied in video surveillance yet it is still a challenging problem. One of the major challenges lies in the lack of spatial and temporal cues, which makes it difficult to deal with large variations of lighting conditions, viewing angles, body poses and occlusions. Recently, several deep learning based person re-identification approaches have been  proposed and achieved remarkable performance. However, most of those approaches extract discriminative features from the whole frame  at one glimpse without differentiating various parts of the persons to identify. It is  essentially important to examine multiple highly discriminative local regions  of the person images in details through multiple glimpses for dealing with the large appearance variance. 

In this paper, we propose a new soft attention based model, \emph{i.e.}, the end-to-end Comparative Attention Network (CAN), specifically tailored for the task of person re-identification. The end-to-end CAN learns to selectively focus on parts of pairs of person images after taking a few glimpses of them and  adaptively \emph{comparing} their appearance.  The CAN model is able to learn which parts of images are relevant for discerning persons and automatically integrates information from  different parts to determine whether a pair of images belongs to the same person. In other words, our proposed CAN model simulates the human perception process to verify whether two images are from the same person. Extensive experiments  on four benchmark person re-identification datasets, including CUHK01, CHUHK03, Market-1501 and VIPeR, clearly demonstrate that our proposed end-to-end CAN for person re-identification outperforms well established baselines significantly and offer new state-of-the-art performance.          
\end{abstract}

\begin{IEEEkeywords}
Person re-identification, Comparative Attention Network, Multiple glimpses.
\end{IEEEkeywords}

%
\IEEEpeerreviewmaketitle

\section{Introduction}
\IEEEPARstart{R}{ecently}, person re-identification (re-id), \emph{i.e.}, person or pedestrian re-identification across multiple cameras without overlapping view, has received increasing attention~\cite{davis2007information,gray2008viewpoint,guillaumin2009you,farenzena2010person,hirzer2012person,mignon2012pcca,koestinger2012large,paisitkriangkrai2015learning,li2013locally,zhao2013unsupervised,zhao2013person,zhao2014learning,li2014deepreid,yi2014deep,zheng2015scalable,ahmed2015improved,wu2016personnet, assari2016re, assari2016human, mclaughlinrecurrent, LeAn:2015bn, an2016person, an2016personRD, liao2015person, zheng2016towards, xiao2016learning, cheng2016person, zhang2016learning, matsukawa2016hierarchical, Zhang_2016_CVPR, liao2015efficient, su2015multi, shen2015person, martinel2016temporal, shi2016embedding, Chen_2016_CVPR, tan2016dense, zheng2016learningbinary }. 
It aims to re-identify a person that has been captured by one camera in another camera at any new location. Person re-identification has many important applications in security systems and video surveillance of public scenarios such as stores and shopping malls. Different from the video-based person re-identification methods, such as \cite{mclaughlinrecurrent, liu2017video}, we investigate the problem of image-based person re-identification in this paper. Therefore, one of the major challenges lies in the lack of spatial and temporal cues. Moreover, it is  also challenging to obtain satisfactory results in terms of accuracy  in real-world scenarios due to the large appearance variance across multiple cameras. For example, people usually pose differently in two different views. Besides, other factors such as variations in color, illumination, occlusion as well as low-resolution of the captured frames also increase  difficulties of the realistic person re-identification.

\begin{figure}[t]
	\centering
	\includegraphics[width=8.5cm]{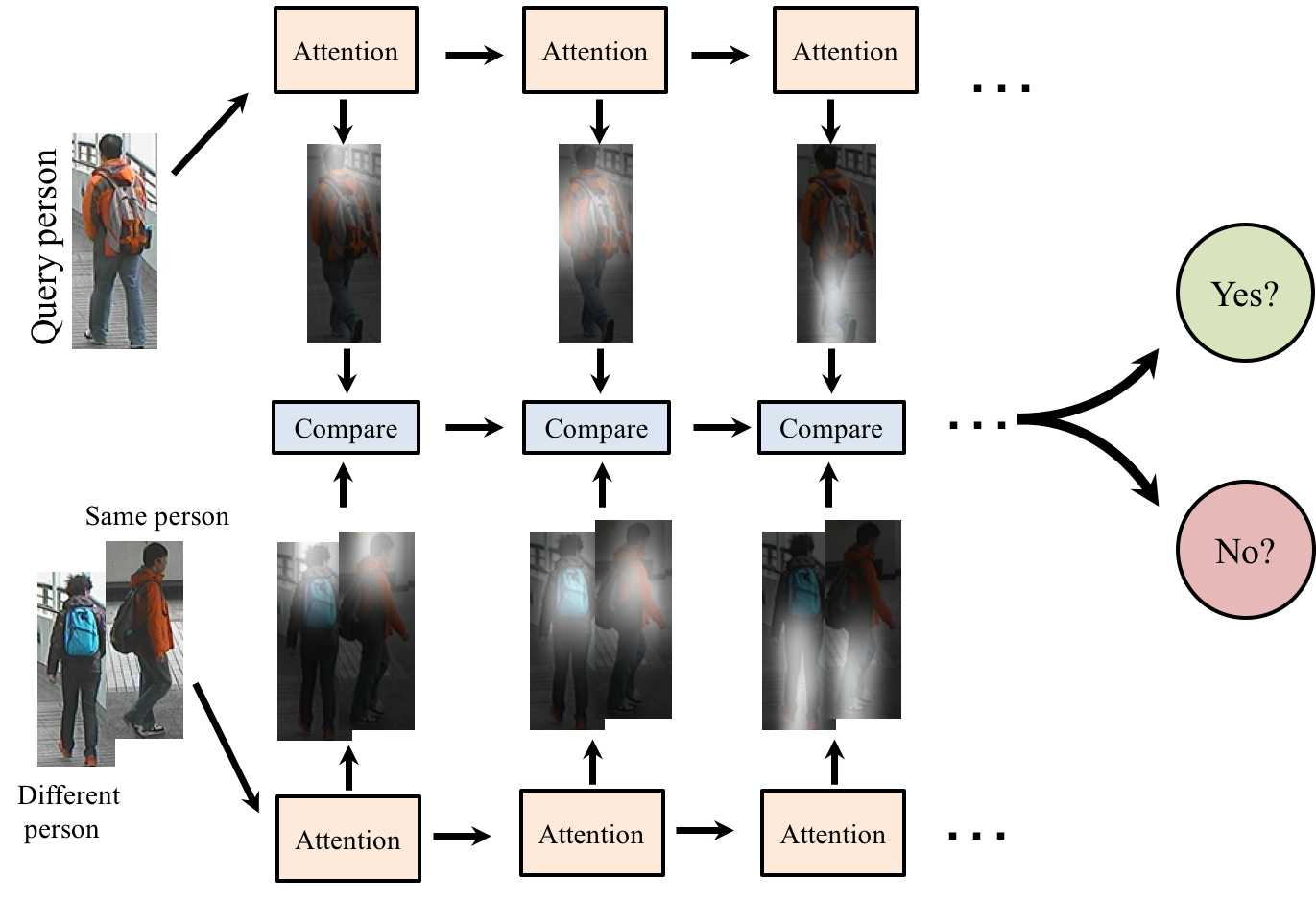}
	
	\caption{Illustration  of motivation of our method. The top image is the image of a query person while the bottom gives two images containing a same and a different person with the query. Through repeatedly comparing person pairs, a series of different local parts (\emph{e.g.,} head parts, torso parts and leg parts) of persons are focused on (highlighted by the white regions), and then the information from the different parts is integrated to discern whether the image pairs belong to the same person.  } 
	
	
	\label{fig:moti}
\end{figure}

Recently, several research attempts have been made for solving the re-identification problem through effectively reasoning over the person appearance. Some approaches for person re-identification \cite{li2014deepreid,yi2014deep,ahmed2015improved,wu2016personnet, cheng2016person, shi2016embedding } utilize the Convolutional Neural Networks (CNNs) to learn effective representation of person appearance and achieve remarkable performance. Some of them~\cite{yi2014deep,ahmed2015improved, cheng2016person, shi2016embedding} exploit part or patch matching-based architecture to learn discriminative feature representation in local regions of persons. However, the parts in most of these methods are all pre-defined. For example, \cite{shi2016embedding} splits the input person image into three square overlapping patches from top to bottom, and applies CNN architecture on them to learn the discriminative features of each patch. The performance  of re-identification may be influenced by the split way of regions.  Additionally, the above methods learn the  local discriminative representation from local regions once for all.  As a consequence, the performance of such approaches may still suffer from factors such as illumination variance and occlusion if the learned local feature representation is not so robust to the factors.  Considering the problems above, we propose a model which can adaptively find \emph{multiple} local regions with more discriminative information in person images in a recurrent way  and integrate them to further improve the person re-identification performance. 

According to related research \cite{davis2007information,gray2008viewpoint,guillaumin2009you,farenzena2010person,hirzer2012person,mignon2012pcca,koestinger2012large,paisitkriangkrai2015learning,li2013locally,zhao2013unsupervised,zhao2013person,zhao2014learning,li2014deepreid,yi2014deep,zheng2015scalable,ahmed2015improved,wu2016personnet} and our daily experience,  in the process of a human discerning another in a crowd,  the human often abstracts the discriminative features of all the individuals and then compares the similarity and difference of them to find the specific one correctly, and this process can be repeated many times (i.e. multiple glimpses of each person). At the end of the process, the information gathered from glimpses is integrated as the comprehensive information to help the discerning. The motivation is illustrated in Fig.\ref{fig:moti}. The focused parts in the repeated comparison process are highlighted by the white regions, corresponding to heads, torsos, and legs, which can provide discriminative information to identify persons. For example, whether they are wearing the same jackets or carrying the same  backpack. Inspired by the observation, we propose an attention based model with inherent comparative components to solve the person re-identification problem.

With the recent development of Recurrent Neural Networks (RNNs) based on Long Short-Term Memory (LSTM)~\cite{hochreiter1997long}, the attention based models have demonstrated outstanding performance on several challenging sequential data recognition and modeling tasks, including caption generation~\cite{xu2015show}, machine translation~\cite{bahdanau2014neural}, as well as action recognition~\cite{sharma2015action}. Briefly, similar to human visual processing, attention-based algorithms tend to selectively concentrate on a part of the information, and at the same time ignore other perceived information. Such a mechanism is usually called \textit{attention} and  can be employed to adaptively localize discriminative parts or regions of person images. Thus it is helpful to solve the person re-identification problem, which however has been rarely considered in the  literatures. 

In this work, we go beyond the standard LSTM based attention models and propose an end-to-end Comparative Attention Network (CAN). 
The proposed end-to-end CAN framework simulates the re-identification process of human visual system by learning a comparative model from raw person images to recurrently localize some discriminative parts of person images via a set of glimpses. At each glimpse, the model generates different parts without any manual annotations. It takes both the raw person images and the locations of a previous glimpse as inputs, and produces the next glimpse local region features as the outputs. These features can be regarded as a kind of dynamical pooling feature, and we show that exploiting these features generated by our CAN model for person re-identification performs better than conventional pooling features, which is used by many existing models~\cite{ahmed2015improved,li2014deepreid,yi2014deep,wu2016personnet}. 

Compared with the work of attention-based action recognition~\cite{sharma2015action} using video sequence, our work is more related to the attention-based image caption generation~\cite{xu2015show} which also applies the attention model on the still images to learn a series of different local attention features in a recurrent way. However, our proposed CAN framework end-to-end learns the attention regions from raw images while the model in \cite{xu2015show} generates attention regions based on the pre-extracted CNN features. Furthermore, our model has a comparing ability in the generation of  local attention regions as it is a three-branch architecture taking a triplet of images as input for each branch while \cite{xu2015show} only has one branch structure taking one image as input. In contrast, our approach is also able to achieve comparatively better performance compared to other methods, as validated by experimental results. 

\begin{figure*}[t]
	\centering
	\includegraphics[width=18cm]{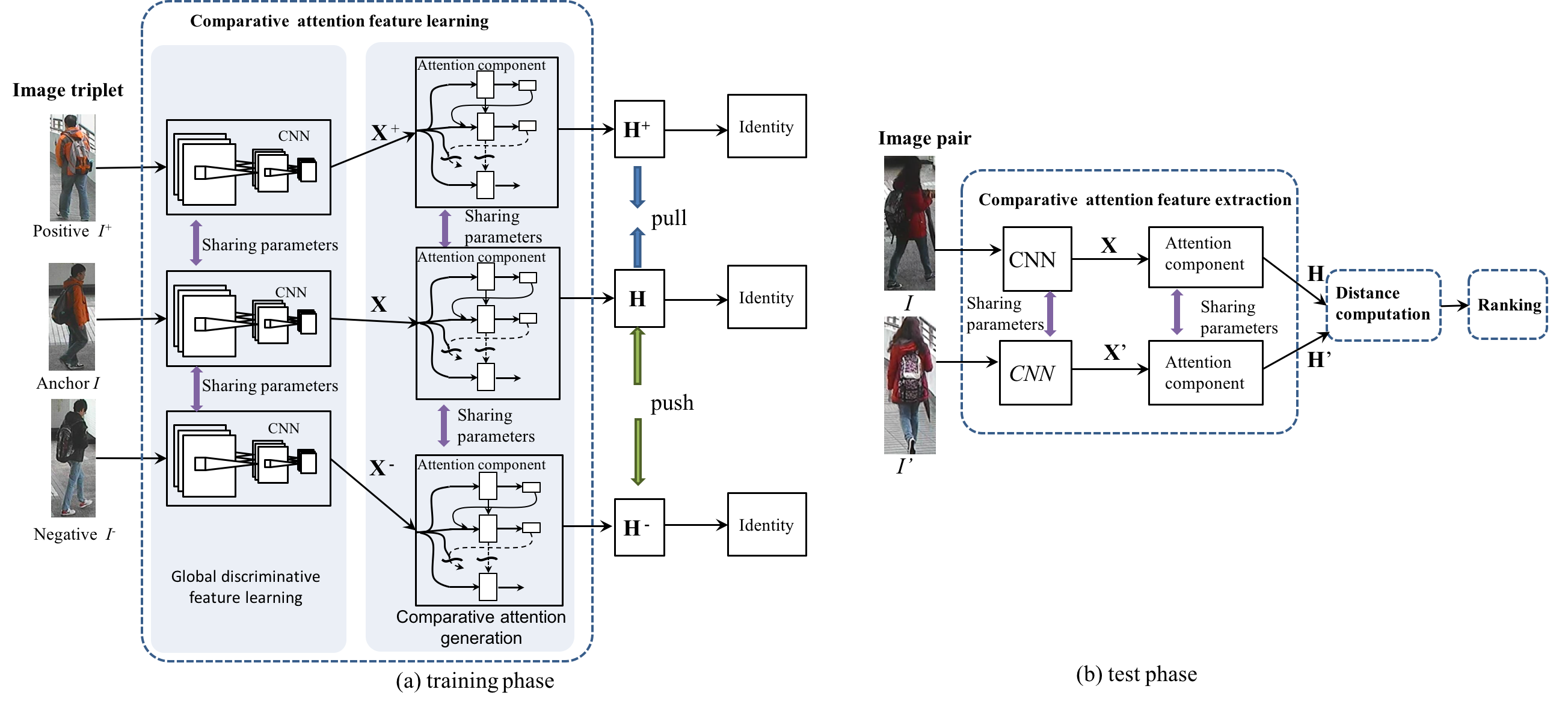}
	
	\caption{The architecture of the proposed Comparative Attention Network (CAN). It consists of two parts, i.e., global discriminative features learning (CNNs) and local comparative visual attention generation. In the training phase, the model utilizes the weight-shared CNN to learn global features from a triplet of images and then passes them to the comparative attention component to compare positive pair and negative pair in a triplet to obtain the discriminative local visual attention features $\mathbf{H}$, $\mathbf{H}^+$ and $\mathbf{H}^-$, which makes positive pairs closer whereas negative pairs further away from each other in each triplet. These comparisons performed among positive pairs and negative pairs are achieved by introducing the step-wise triplet loss and  identification loss on top of the network (refer to text for more details). And the whole architecture is trainable in an end-to-end way. In the test phase, the trained model is applied to each pair of persons and distances between each pair are computed for ranking the pairs of query and candidate frames. 
	}
	
	\label{fig:train_arch}
\end{figure*}

In summary, we make following  contributions to person re-identification: 
\begin{itemize} 
	\item We propose a new attention model that dynamically generates discriminative  features in a recurrent way of ``seeing'' and ``comparing'' person images  for automatically localizing the most discriminative parts of persons.
	\item We develop a comparative network that can efficiently seek discriminative parts of person image pairs by incorporating an on-line triplet selection method. Moreover, our CAN framework is able to generate attention parts directly from raw person image pairs in an end-to-end way. 
	\item Finally, we quantitatively validate the good performance of
	our end-to-end CAN framework by comparing it to the state-of-the-art
	performance on four benchmark datasets: CUHK01 \cite{li2013locally}, CUHK03 \cite{li2014deepreid}, Market-1501 \cite{zheng2015scalable} and VIPeR \cite{gray2007evaluating}. 
\end{itemize}

The paper is organized as follows. Sec. II reviews the related work briefly. In Sec. III, the framework is described in details. Then, the experimental results on several public benchmark datasets are shown and the analyses are given in Sec. IV. Finally, a conclusion is presented in Sec. V.

\section{Related Work}
Typically, extracting features from input images and seeking a metric for comparing these features across images are two main components of person re-identification. The
basic thought of searching for better feature representation is to find features that are partially invariant to lighting, pose, and
viewpoint changes. A part of  existing methods primarily employ hand crafted features such as color and texture histograms. Some studies have obtained more discriminative and robust feature representation, such as Symmetry-Driven Accumulation of Local Features (SDALF) \cite{farenzena2010person} exploiting both symmetry and asymmetry color and texure information. Local Maximal Occurrence (LOMO)~\cite{liao2015person} analyzes the horizontal occurrence of local features, and maximizes the occurrence to make a stable representation against viewpoint changes.  In \cite{an2016personRD}, the authors proposed the reference descriptors (RDs) generated with the reference set to improve the matching rate. To utilize complementary information from different feature descriptors, a multiple hypergraph fusion (multi-HG) method was proposed in \cite{an2016person} to learn multiple feature descriptors. In \cite{tan2016dense}, a ranking method fusing the dense invariant features (DIF) was proposed to model the relationship between an image pair across different camera views.  

Additionally, some saliency-related methods \cite{zhao2013unsupervised}, \cite{zhao2013person} have been proposed to enhance the ability of representation and discrimination of the feature for person re-identification. In \cite{zhao2013unsupervised}, the authors presented a method of adjacency constrained patch matching to build dense correspondence between image pairs in an unsupervised way. Moreover, an approach called SalMatch~\cite{zhao2013person} integrated both salience matching and patch matching based on the RankSVM framework. And mid-level filters (MidLevel)~\cite{zhao2014learning} was learned from patch clusters with coherent appearance obtained by pruning hierarchical clustering trees to get view-invariant and discriminative features. However, the above methods pre-extract the low-level or mid-level features of pre-defined local regions and then generate saliency maps. The feature extraction and  saliency map generation are two separate processes, which would  affect the person re-identification performance. Compared with the mentioned saliency-based methods, our attention-based CAN model is also a kind of saliency method. However, it is able to automatically learn the attention maps from raw person images in an end-to-end way.

Compared with the methods where complicated hand-crafted feature representation is designed, there are some approaches using metric learning, which formulate the person re-identification as a supervised distance metric learning problem where a transformation matrix is learned so that the Mahalanobis distance is relatively small when extracted features represent a same person and big otherwise. To achieve this goal, Large Margin Nearest Neighbor   (LMNN)~\cite{hirzer2012person}, Logistic Discriminant Metric Learning  (LDM)~\cite{guillaumin2009you}, Information Theoretic Metric Learning (ITML)~\cite{davis2007information}, Kernelized Relaxed Margin Components Analysis (KRMCA)~\cite{liu2015kernelized}, Robust Canonical Correlation Analysis (ROCCA)\cite{LeAn:2015bn},  Metric Learning with Accelerated Proximal Gradient (MLAPG)\cite{liao2015efficient} and transfer Local Relative Distance Comparison (t-LRDC) are representative methods. Inspired by the thought of comparison in \cite{hirzer2012person}, \cite{guillaumin2009you}, \cite{davis2007information} and \cite{liu2015kernelized}, we exploit the multi-task loss including triplet loss and identification loss in this paper,  forcing the similarities of instances ``more similar" and diversities ``more different". Besides, in \cite {mignon2012pcca}, Pairwise Constrained Component Analysis (PCCA) learns distance metrics from sparse pairwise similarity/dissimilarity constraints in the high dimensional input space. And large scale metric learning from equivalence constraint  (KISSME)~\cite{koestinger2012large} considers a log-likelihood ratio test of two Gaussian distributions. In \cite{zheng2016towards}, a transfer local relative distance comparison (t-LRDC) model was formulated to address the open-world person re-identification problem by one-shot group-based verification.To match people from different views in a coherent subspace, \cite{LeAn:2015bn} proposed a robust canonical correlation analysis (ROCCA) method. And in \cite{liao2015efficient}, the Metric Learning with Accelerated Proximal Gradient (MLAPG) was proposed to solve the person re-identification problem. In \cite{paisitkriangkrai2015learning}, the authors presented an effective structured learning based approach by combining multiple low-level hand-crafted and high-level visual features. 

Despite the hand-crafted features based methods aforementioned, there are several deep learning based person re-identification approaches proposed~\cite{ahmed2015improved, li2014deepreid, yi2014deep, wu2016personnet, cheng2016person, shi2016embedding}. \cite{li2014deepreid} proposed a method learning a Filter Pairing Neural Network (FPNN) to encode and model photometric transforms by using the patch matching layers to match the filter responses of local across-view patches for person re-identification. In~\cite{yi2014deep}, a Siamese CNN, which is connected by a cosine layer, jointly learns the color feature, texture feature and metric in a unified framework. Moreover, \cite{wu2016personnet} improved the performance by increasing the depth of layers and using very small  convolution filters. In \cite{cheng2016person,  shi2016embedding}, the authors  proposed parts-based CNN model to learn the discriminative representations. Different from the part-based CNN methods~\cite{yi2014deep,ahmed2015improved, cheng2016person, shi2016embedding}, our method is able to learn the local discriminative regions rather than pre-defining or splitting the local parts.

Recently, LSTMs have shown good performance in the domain of speech recognition~\cite{graves2013hybrid} and image description \cite{xu2015show}. More recently, \cite{sharma2015action}  developed recurrent soft attention based models for action recognition and analysed where they focus their attention. Especially, the authors suggested that it is quite difficult to interpret internal representations learned by deep neural networks. Therefore, attention models add a dimension of interpretability by capturing where the model is focusing its attention when performing a particular task. Inspired by the above work, in this paper, we employ the recurrent attention model to generate different attention location information by comparing image pairs of persons and then integrate them together. As far as we know, our work is the first one applying the attention model to the person re-identification problem. Similar to saliency-related \cite{zhao2013unsupervised}, \cite{zhao2013person}, \cite{zhao2014learning} methods mentioned above, the attention model is also a kind of saliency to certain extent, but our attention model can directly obtain the saliency-like attention maps from raw person image due to the end-to-end training pattern. Moreover, different from other attention models, our attention model generates attention maps based on the comparison over image triplets of people. Consequently, our network outperforms all previous approaches on benchmark person re-identification datasets.

\section{Model Architecture}\label{mdac}
In this paper, we propose an end-to-end Comparative Attention Network (CAN) based architecture that formulates the problem of person re-identification as discriminative visual attention finding and ranking optimization. Fig. \ref{fig:train_arch} illustrates our network architecture (\ref{net_arch}). For a given triplet of raw person images, we apply end-to-end Comparative Attention Network (CAN) at each one to learn comparative attention features. The global discriminative features are learned by CNNs, and then passed to the LSTM-based (\ref{lstm_can}) comparative attention components (\ref{cac}) to obtain the discriminative attention masked features at different time steps. 
To combine these different time step features and make them more discriminative, a triplet selection  method (\ref{trip_sel}) is utilized after concatenating different time step features. Each of these components is explained in the following subsections.

\subsection{End-to-End Comparative Attention Network Architecture}\label{net_arch}

Fig. \ref{fig:train_arch} illustrates the architecture of the proposed end-to-end Comparative Attention Network (CAN). The CAN network can localize and compare multiple person  parts using the comparative attention mechanism. In this section, we describe how our comparative attention network works in the training phase and the test phase individually.

\subsubsection{Training Phase}
During training, the model starts from processing a triplet of raw images. Here, we denote the images of a triplet as $I$, $I^+$ and $I^-$, corresponding to the anchor sample, the positive sample and the negative sample respectively. $I$ and $I^+$ come from the same class (positive pair), while $I^-$ is from a different class (negative pair). The objective of CAN is to learn effective feature representation and to generate discriminative visual attention regions. Thus, in terms of the features extracted from the attention regions,  the truly matched images are closer than the mismatched images by training the model on a set of triplets $\langle I,I^+,I^-\rangle$. Fig. \ref{fig:train_arch} (a) shows the overall  architecture used for training. The comparative attention network consists of following two parts: global discriminative feature learning components and comparative attention components.

\noindent{\textbf{Comparative Attention Feature Learning:\quad}}
In this paper, we adopt the truncated CNN such as Alexnet\cite{krizhevsky2012imagenet} and VGG~\cite{simonyan2014very} for global discriminative feature learning, and the learned feature map is denoted as  $\mathbf{X}=\phi_{CNN}(I)$. Before end-to-end training the whole CAN, we pre-train the CNN with softmax classification model, which contains several convolutional feature learning layers with three fully-connected classification layers followed.  After the pre-training of this network, the last three fully-connected layers are replaced with our proposed comparative attention model. The truncated CNN are used to learn global discriminative appearance features. Then, they are passed to the comparative attention components our proposed CAN to generate the comparative visual attention regional features, which are denoted as $\mathbf{H}=\beta(\mathbf{X})$. Here, $\beta$ denotes the comparative attention generation part of our model, and $\mathbf{H}$ correspond to local comparative attention features of persons. Note that all the person samples in a triplet share the same parameters in feature learning and comparison, as shown in Fig. \ref{fig:train_arch} (a). Details of the comparative attention model will be given in  Section \ref{lstm_can} and \ref{cac}.

\noindent {\textbf{Multi-task Loss:\quad} As mentioned above, our goal is to learn discriminative feature representation and visual attention regions through comparing the similarity and difference of positive and negative pairs in each triplet. Therefore, similar to \cite{sun2014deep, mclaughlinrecurrent}, we adopt the multi-task loss including triplet loss \cite{schroff2015facenet} and identification loss  as the final loss function. 

 Within a triplet of $\langle
 \mathbf{H}_n, \mathbf{H}_n^+, \mathbf{H}_n^-\rangle$, we expect  features of the positive sample $\mathbf{H}_{n}^+$ is more similar to $\mathbf{H}_n$ than the features of the negative sample: 
\begin{equation}\label{eq1}
\left \| \mathbf{H}_{n}-\mathbf{H}_{n}^{+} \right \|_{2}^{2}+\alpha < \left \| \mathbf{H}_{n}-\mathbf{H}_{n}^{-} \right \|_{2}^{2}.
\end{equation}
Here $\alpha$ is a margin that is introduced to enhance the discriminative ability of learned features  between positive and negative pairs. Therefore, for $N$ triplets, one loss function that CAN is going to minimize is: 
\begin{equation}\label{trip_loss}
\mathcal{L}_{trip}=\frac{1}{N}\sum_{n}^{N}\left[\left \| \mathbf{H}_{n}-\mathbf{H}_{n}^{+} \right \|_{2}^{2}- \left \| \mathbf{H}_{n}-\mathbf{H}_{n}^{-} \right \|_{2}^{2}+\alpha\right]_{+},
\end{equation}
where $[\cdot]_+$ truncates the involved variable at zero.

Additionally, for each $\delta$-dimension feature vector $\mathbf{H}$ output by our CAN, the identity of  the person is predicted using the standard softmax function, which is defined as follows: 
\begin{equation}
\Omega(\mathbf{H}) = P(z=v|\mathbf{H})=\frac{\exp(S_{v}^\top\mathbf{H})}{\sum_{g}\exp(S_{g}^\top \mathbf{H})},
\end{equation}
where \textit{G} is the total number of identities, $z$ is the predicted identity for the input person, and $S \in \mathbb{R}^{ \delta \times G }$ is the weight matrix used in the softmax function, and  $S_{v} \in \mathbb{R}^{\delta }$ and $S_{g} \in \mathbb{R}^{\delta} $ represent the \textit{v}$^{th}$ and $g^{th}$ column of it, respectively. Then, for \textit{N} triplets,  the corresponding softmax loss function is defined as follows: 
\begin{equation}\small
\mathcal{L}_{iden}= \frac{1}{3N}\sum_{n}^{N}\left( -\textup{log}(\Omega(\mathbf{H}_{n}))-\textup{log}(\Omega(\mathbf{H}_{n}^{+}))-\textup{log}(\Omega(\mathbf{H}_{n}^{-}))\right).
\end{equation}

Finally, we jointly end-to-end train our architecture with both triplet loss and identification loss.  We can now define the overall multi-task training loss function  $\mathcal{L}_{multi}$  which jointly optimizes the  triplet cost and the identification cost as follows:
\begin{align}
\mathcal{L}_{multi}( \mathbf{H}_n, \mathbf{H}_n^+ + \mathbf{H}_n^-) = &\mathcal{L}_{trip}( \mathbf{H}_n, \mathbf{H}_n^+ + \mathbf{H}_n^-)\notag\\ &+ \mathcal{L}_{iden}( \mathbf{H}_n, \mathbf{H}_n^+, \mathbf{H}_n^-).
\end{align}
Here, we give equal weights for the triplet cost and identification cost terms. The above comparative attention network (including CNNs and attention components) can be trained end-to-end using back-propagation from raw person images (details of our training parameters can be found in Sec. \ref{para_set}). Next, we proceed to introduce how to apply the network trained for testing.

\subsubsection{Test Phase}
As shown in Fig.~\ref{fig:train_arch} (b), in the test phase, we pass a set of person image pairs in the testing set into the trained CAN, where the Euclidean distance of them is computed. Then the ranking unit directly outputs the final ranking results. Here, we adopt average CMC (Cumulative Matching Characteristics) \cite{gray2007evaluating} and the accuracy at top ranks as the evaluation metrics, as in \cite{davis2007information,gray2008viewpoint,guillaumin2009you,farenzena2010person,hirzer2012person,mignon2012pcca,koestinger2012large,paisitkriangkrai2015learning,li2013locally,zhao2013unsupervised,zhao2013person,zhao2014learning,li2014deepreid,yi2014deep,zheng2015scalable,ahmed2015improved,wu2016personnet}. The detailed definition of CMC will be given in the Sec.~\ref{eval_prot}.  It is worth to mention that we also examine these items to see the performance of the whole network on the validation dataset during the training phase. This is because the training loss can only reflect the tendency of performance variance on the training set while the output evaluations on the validation set can directly indicate the true ranking performance. That is, we can train the network through directly optimizing the ranking results on the validation set.
\begin{figure}[htb]
	\centering
	\includegraphics[width=8cm]{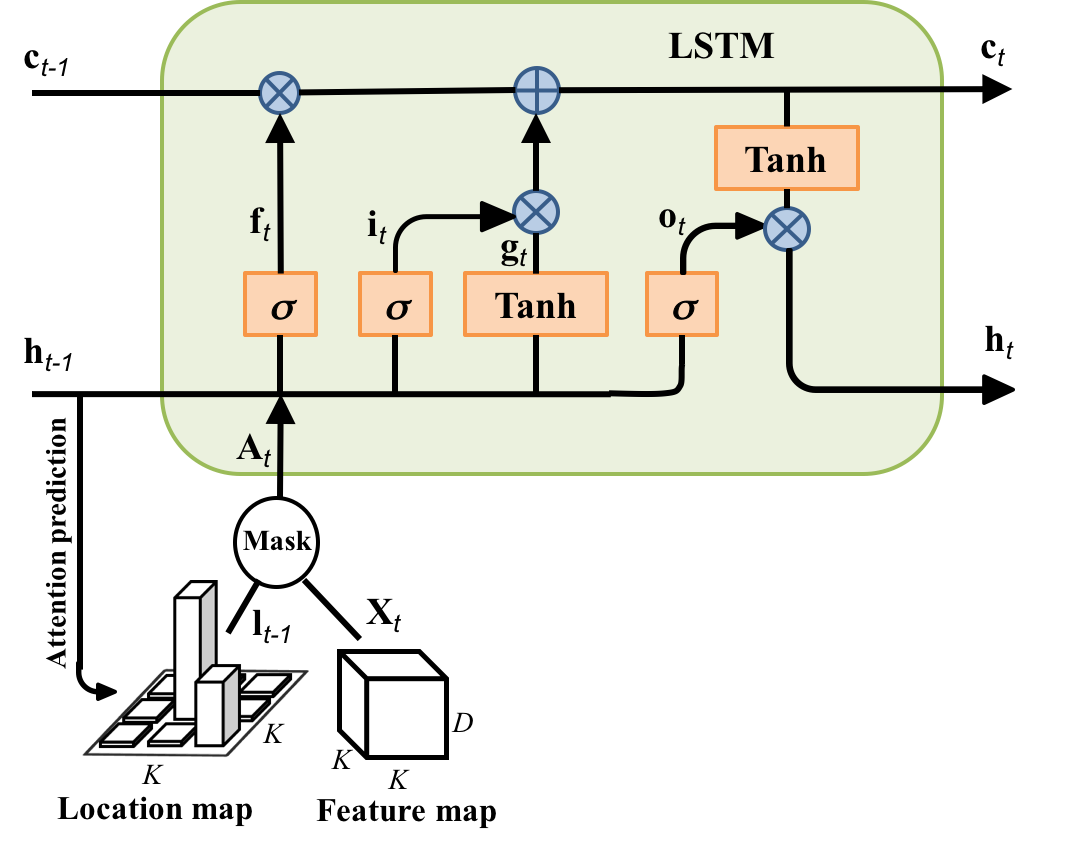}
	\caption{A single time step LSTM unit which takes masked CNN feature map $\mathbf{A}_t$ and previous hidden state $\mathbf{h}_{t-1}$ as input. The location map or attention map $\mathbf{l}_{t-1}$ masking on the input feature map $\mathbf{X}_t$ is predicted by the previous hidden state $\mathbf{h}_{t-1}$.  Each component of LSTM learns how to cooperate to weigh the input information (input gate $\mathbf{i}_t$ ), \textit{i.e.} to remember the useful one (memory state $\mathbf{c}_t$) and to erase the unnecessary one (forget gate $\mathbf{f}_t$). Finally, the output gate $\mathbf{o}_t$ controls how the filtered memory should be emitted.      
	}
	\label{fig:LSTM}
\end{figure}         
\subsection{ Long Short-Term Memory Networks } \label{lstm_can}

In our CAN model, we use a long short-term memory (LSTM) network to produce an attention map over a local region at every time step conditioned on the input CNN feature maps, the previous hidden states and the generated attention map in the previous step. We implement the LSTM by following \cite{zaremba2014recurrent}, \cite{xu2015show} and \cite{sharma2015action}, which is also illustrated in Fig.~\ref{fig:LSTM}. At the time step $t$, LSTM takes a masked CNN feature map $\mathbf{A}_t$ and the previous hidden state $\mathbf{h}_{t-1}$ as inputs. The attention map $\mathbf{l}_{t-1}$ is predicted from the previous hidden state $\mathbf{h}_{t-1}$ using the learned parameters $W_{i,h}$. The predicted attention map is then used to mask the input feature map $\mathbf{X}_t$ of size $K \times K \times D$, giving a filtered feature maps where only attended regions are preserved. The formulations are shown as follows: 
\begin{align}
\mathbf{i}_{t} &=\sigma \left( \mathbf{M}_{i} \left[\mathbf{h}_{t-1}, \mathbf{A}_{t}\right] + b_i\right), \nonumber\\
\mathbf{f}_{t} &=\sigma \left( \mathbf{M}_{f} \left[\mathbf{h}_{t-1}, \mathbf{A}_{t}\right] + b_f\right),\nonumber\\
\mathbf{o}_{t} &=\sigma \left( \mathbf{M}_{o} \left[\mathbf{h}_{t-1}, \mathbf{A}_{t}\right] + b_o\right),\\
\mathbf{g}_{t} &=\textrm{tanh} \left( \mathbf{M}_{g} \left[\mathbf{h}_{t-1}, \mathbf{A}_{t}\right] + b_g\right),\nonumber\\
\mathbf{c}_t &= \mathbf{f}_t \odot \mathbf{c}_{t-1} + \mathbf{i}_t \odot \mathbf{g}_t,\nonumber\\
\mathbf{h}_t &= \mathbf{o}_t \odot \textrm{tanh}\left(\mathbf{c}_t\right),\nonumber
\end{align}
where $\mathbf{i}_{t},\mathbf{f}_{t},\mathbf{c}_{t},\mathbf{o}_{t}$ and $\mathbf{h}_{t}$ are the input gate, forget gate, cell state, output gate and hidden state\cite{hochreiter1997long} respectively. $\mathbf{M}_{\sim}$ ($\mathbf{M}_{\sim}=\left\lbrace  \mathbf{M}_{i}, \mathbf{M}_{f}, \mathbf{M}_{o}, \mathbf{M}_{g} \right\rbrace $) and $b_{\sim}$  ($b_{\sim}=\left\lbrace b_i, b_f, b_o, b_g  \right\rbrace $) denote learnable weight parameters inside the gates and $\odot$ is the Hadamard product. 

To produce the attention map,  at each time step $t$, the softmax location map (\textit{i.e.}, the attention map) $\mathbf{l}_{t-1}$ of size $K \times K$ is predicted from the previous hidden state $\mathbf{h}_{t-1}$ by the learnable parameters $W_{i,h}$ as follows:
\begin{equation}
l_{t-1,i}  = \frac{\exp(W_{i,h}^\top \mathbf{h}_{t-1})}{\sum_{j=1}^{K\times K}\exp(W_{j,h}^\top \mathbf{h}_{t-1} )}, \quad i = 1, \ldots, K^2,
\end{equation}
where the weight parameters  $W_{i,h}$ to generate softmax locations (attention) are learned together with the gate parameters $\mathbf{M}_{\sim}$ and $b_{\sim}$ in the end-to-end model training.  Then, the masked feature $\mathbf{A}_t$ (through weighted average pooling) produced at the time step $t$  is computed as follows:
\begin{equation}
 \mathbf{A}_t = \mathbb{E}_{p(\mathbf{l}_{t-1}|\mathbf{h}_{t-1})}[\mathbf{X}_t]=\sum_{i=1}^{K^2}l_{t-1,i}
 \mathbf{X}_{t,i},
\end{equation}  
where $\mathbf{X}_{t}$ and $\mathbf{X}_{t,i}$ correspond to the CNN feature maps and the $i^{th}$ slice of the feature cube at time step $t$. Here $\mathbf{X}_{t}$ is a tensor of $K \times K \times D$. Each location out of $K \times K$ locations is described by a $D$-dimensional feature. The dimension of $\mathbf{A}_{t}$ is $1 \times 1 \times D$. Note, for each time-step $t$, our model takes the same CNN feature map $\mathbf{X}$  as input, so the CNN features $\mathbf{X}_{t}$ are the same for all the time steps. We adopt the following initialization method for memory state and hidden state:
\begin{align}
\mathbf{c}_0=f_{\text{init},c}\left(\frac{1}{K^2}\sum^{K^2}_{i=1}\mathbf{X}_{0,i}\right),\label{c_ini}\\
\mathbf{h}_0=f_{\text{init},h}\left(\frac{1}{K^2}\sum^{K^2}_{i=1}\mathbf{X}_{0,i}\right),\label{h_ini}
\end{align} 
where $f_{\text{init},c}$ and $f_{\text{init},h}$ are two-layer perceptrons consisting of two Fully Connected (FC) layers which can be learned end-to-end with other model components. And $\mathbf{X}_{0,i}$ represents the $i^{th}$ slice of the feature map $\mathbf{X}_{0}$  output by CNN (corresponding to $\mathbf{X}_{0}$ in Fig. \ref{fig:att_cp}). These values are used to calculate the initial softmax attention location $\mathbf{l}_0$ which is applied on CNN features $\mathbf{X}_1$ to get the initial input $\mathbf{A}_1$ as shown in Fig. \ref{fig:LSTM} and Fig. \ref{fig:att_cp}. Note that different from~\cite{sharma2015action}, our CAN model is trained end-to-end, so the  CNN is trained together with the attention model  of our CAN model in the training process, while in \cite{sharma2015action} the CNN features are pre-extracted off-line to initialize the memory state and hidden state  for the attention model.

\begin{figure}[htb]
	\centering
	\includegraphics[width=8.5cm]{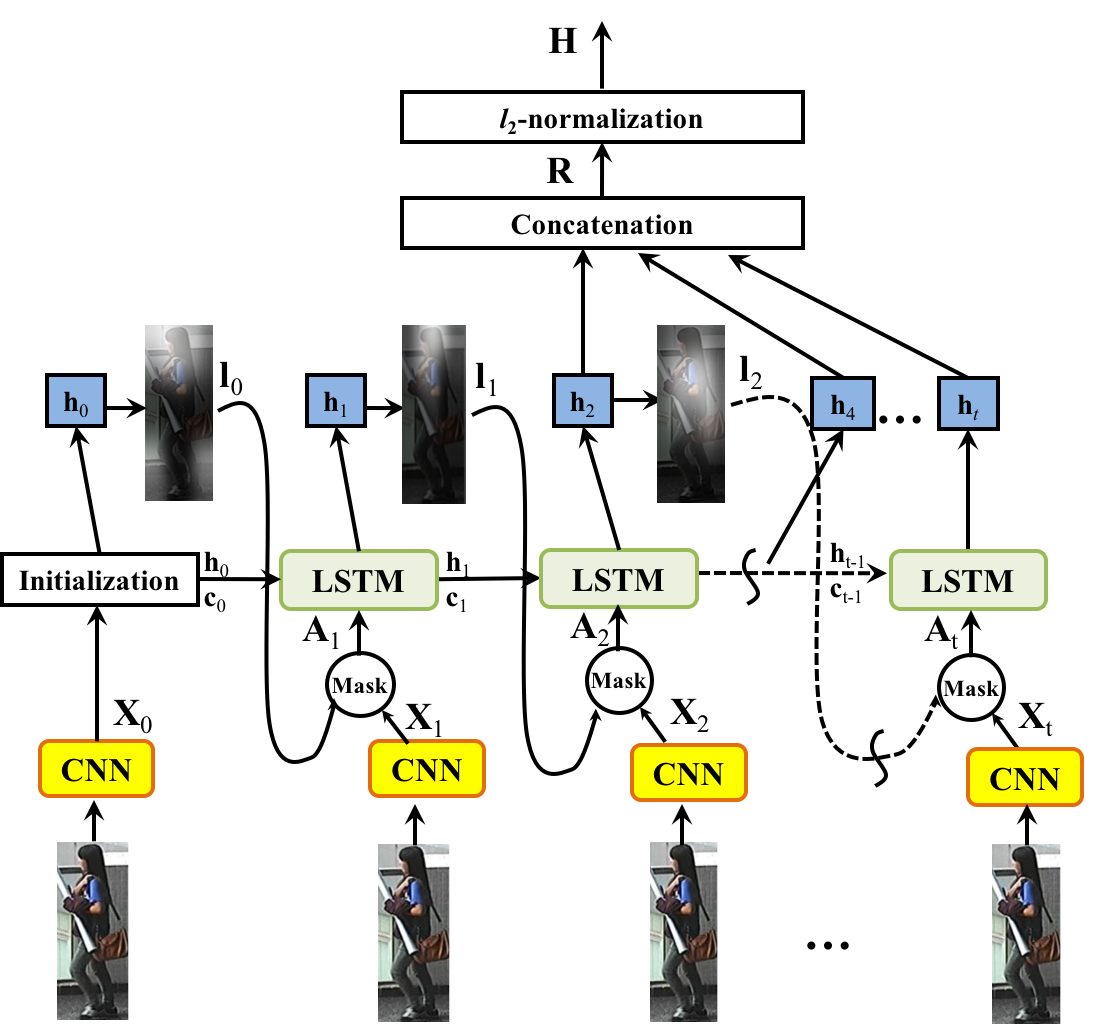}
	\caption{The comparative attention component composed of several time step LSTMs takes CNN feature maps of a same person image as its input at each time step and  outputs the concatenated hidden states, which are utilized every several time steps, as the features sent to the triplet loss layer. Here, we simply show one attention component, but there still exist another two weight-sharing attention components working simultaneously to compare positive and negative pairs in triplets and generate the comparative attention maps in training phase.
	}
	
	\label{fig:att_cp}
\end{figure}     
  
 \subsection{The Comparative Attention Component}\label{cac}
 In our comparative attention network, the LSTM-based comparative attention components take the feature maps output by CNN as inputs. Generally, CNN is used to obtain the global discriminative features from raw person images while the comparative attention components generate more local attention regions through comparing different person images. Intuitively, this mechanism has somewhat similarity with the function of the human visual system. In this subsection, we look inside the comparative attention components. The detailed composition  shown in Fig. \ref{fig:LSTM} corresponds to the attention components in the training phase and the test phase in Fig. \ref{fig:train_arch}. The brief working process of the comparative attention components is illustrated in Fig.~\ref{fig:att_cp}. Note that although we simply show one attention component here, there exist three weight-sharing attention components working simultaneously in the training phase, among which two work in the test phase in practice, as shown in Fig.~\ref{fig:train_arch}. At each time step, the attention component takes CNN features learned from a triplet of raw person images as its inputs. Each triplet of person images is ``seen" once by our attention component in one glimpse.

Then some information is ``remembered" and some ``forgotten", decided by the LSTM unit, in order to generate attention location maps and hidden states for the next time step or ``glimpse". For the hidden states output by each time step of  attention components,  they  contain the memory in the process of comparing person images, and are exploited to obtain the local attention maps, as introduced in~\ref{lstm_can}. Therefore, to combine all the generated attention parts information and utilize it as holistic discriminative features for comparison, a concatenation layer is applied to concatenate a few time steps of hidden states along channel axis. The concatenation layer can be defined as follows:
\begin{equation}\label{concat}
\mathbf{R}=[\mathbf{h}_ {\omega_1}; \mathbf{h}_ {\omega_2}; ...; \mathbf{h}_{\omega_m}], \omega_m \in [1, 2, ..., t],
\end{equation}
where $\mathbf{h}_{\omega_m} \in \mathbb{R}^{q \times 1 }$ represents the hidden state with $q$ channels from different time steps.  $t$ is the number of all time steps while $m$ is the number of chosen time steps. We only use hidden steps at $m$ chosen steps to form the matrix $\mathbf{R}$. And $\mathbf{R} \in \mathbb{R}^{mq \times 1 }$ is the $mq$-channel output of the concatenation layer. The concatenation layer plays a crucial role in our CAN model because it integrates different time step features produced by repeatedly comparing person samples so that our model can judge which channels of features from which time step are more important to discerning persons. Through this way, our CAN model can attend different local regions at different time steps. In the Sec. \ref{ana_mod}, we will investigate the effect of choosing features from different time steps to concatenate. 

Additionally, our CAN framework employs the triplet loss as one loss of our multi-mask loss and the whole framework is a three-weights-sharing-branch  as well as recurrent network, so the loss value would fluctuate wildly in the training phase. To ensure that the distance derived from each triplet should not easily exceed the margin $\alpha$ so that more triplet constraints can take effect for the triplet loss function (Eqn. \eqref{trip_loss}), the concatenated features $\mathbf{R}$ are passed to the $\ell_2$-normalization layer as the output of attention components:
\begin{equation}
\mathbf{H}=\frac{\mathbf{R}}{\sqrt{\sum^{mq}_{d=1}R_d^2}},
\end{equation}
where $R_{d}$ represents the  $d^{th}$ entry of $\mathbf{R}$ and $\mathbf{H}$ is also the $mq$-dimension vector.  

As aforementioned, our attention-based CAN model is a kind of saliency method. However, compared with the previous saliency-based methods extracting the saliency of pre-defined regions of persons based on low-level or mid-level features, such as eSDC~\cite{zhao2013unsupervised}, SalMatch~\cite{zhao2013person} and MidLevel~\cite{zhao2014learning}, our method can automatically learn the attention maps from raw person images in an end-to-end way. And the performance of the previous saliency-based methods would be affected by the low-level features because the low-level feature and saliency maps generation are two separate processes. Besides, the previous saliency-based methods get only one saliency map for each person image while our method can learn multiple attention maps which highlight different local regions in a recurrent style. And the superiority of our method is validated in Sec. \ref{cmp_state}.

 \subsection{Triplet Selection} \label{trip_sel} 
 
 It is crucial to select triplets that violate the constraint given in Eqn. \eqref{eq1}. In particular, given $\mathbf{H}_n$, we want to select a positive sample $\mathbf{H}_{n}^{+}$ satisfying $\textup{argmax}_{\mathbf{H}_{n}^{+}}\left \| \mathbf{H}_{n}-\mathbf{H}_{n}^{+} \right \|_{2}^{2}$ while a negative sample satisfying $\textup{argmin}_{\mathbf{H}_{n}^{-}}\left \| \mathbf{H}_{n}-\mathbf{H}_{n}^{-} \right \|_{2}^{2}$. However, it is difficult and unrealistic to compute the argmin and argmax for the whole training set. Furthermore, our model needs to compare pair-wise images and to generate a series of attention locations for every image of each person. Therefore, it requires an efficient way to compute the argmin and argmax. There are two methods to be chosen as mentioned in \cite{schroff2015facenet}:
 \begin{itemize}
 	\item  Off-line triplets selection. The triplets are generated every few steps, and the most recent network checkpoint is employed to compute the argmin and argmax.
 	\item  On-line triplets selection. The selection can be done within a mini-batch. 
 \end{itemize} 
 Obviously, generating all possible triplets would result in overwhelming many
 triplets that are feasible for the constraint
 in Eqn.\ref{eq1}. But some of these triplets would not contribute to the training and slow down the convergence of model training. Besides, they would still
 be passed through the network, which cause large unnecessary resource consumption. Different from off-line triplets selection method, on-line triplets selection approach selects triplets that are active and can contribute to improving the model within a mini-batch, so it is of higher efficiency and lower resource consumption. Therefore, we adopt the on-line triplets selection method in this paper. Specifically, instead of picking the hard positive, we adopt all positive pairs and randomly sample negative samples added to each mini-batch. In practice, we find that using all positive pairs makes the model more stable and converge faster than selectively using hard positive pairs in a mini-batch.
     
\section{Experiments}

\subsection{Datasets and Evaluation Protocol}\label{eval_prot}
There exist several challenging benchmark data sets for person re-identification. In this paper, we use CUHK01~\cite{li2013locally}, CUHK03~\cite{li2014deepreid} Market-1501~\cite{zheng2015scalable}, and VIPeR~\cite{gray2007evaluating}, which are four public benchmarks available, to conduct experiments. In experiments, for each pedestrian, the matching of his or her probe image (captured by one camera) with the gallery images (captured by another camera) is ranked. To reflect the statistics of the ranks of true matches, the Cumulative Match Characteristic (CMC) curve is adopted as the evaluation metric.  Specifically, to create a CMC curve, the Euclidean distances between probe samples and those of gallery samples are computed firstly. Secondly, for each sample, a rank order of all the samples in the gallery is sorted from the sample with the smallest distance to the biggest distance. In the end, the percentage of true matches founded among the first $m$ ranked samples is computed and denoted as rank$m$. Note, all the CMC curves for CUHK01, CUHK03 and VIPeR datasets are computed with single-shot setting. And the experiments on the Market-1501 dataset are under both the single-query and multi-query evaluation settings. In addition, for the Market-1501 dataset, the mean average precision (mAP) as in~\cite{zheng2015scalable}  is also employed to evaluate the performance since there are on average 14.8 cross-camera ground truth matches for each query. To construct the validation set, for the CUHK03 dataset, 100 persons are extracted as the validation set with the similar setting of \cite{li2014deepreid}. For the CUHK01, VIPeR and Market-1501 datasets, the five-fold cross-validation is applied on the training set of each dataset, one as the validation set and the other four as the training set. And the samples in the validation set have no overlap with training set and testing set.
 
\subsubsection{CUHK01}
The CUHK01~\cite{li2013locally} dataset contains 971 persons captured from two camera views, and each of them has two images in each camera view. Camera A captures the individuals in frontal or back views while camera B captures them in side views. 

 We randomly divide the dataset under two settings. The first setting contains  a training set of 871 people and a test set of 100 people. In the second setting, 485 persons are randomly extracted for training while the left 486 persons compose the test set. For each setting, the training/test split is repeated 10 times and the average of CMC curves is reported. 

\subsubsection{CUHK03}
There are 13,164 images of 1,360 identities contained in the CUHK03  dataset~\cite{li2014deepreid}. All pedestrians are captured by six cameras, and each person is only taken from two camera views. It consists of manually cropped person images and images automatically detected by the Deformable-Part-Model (DPM) detector~\cite{felzenszwalb2010object}. This is a more realistic setting considering the existence of misalignment, occlusions, body part missing and detector errors. We evaluate the performance of CAN  with the similar setting of \cite{li2014deepreid}. That is, the dataset is partitioned into three parts: 1,160 persons for training, 100 person for validation  and 100 persons for testing. The experiments are conducted with 20 random splits for computing averaged performance.

\subsubsection{VIPeR}
The VIPeR dataset\cite{gray2007evaluating} is considered the most challenging person re-identification dataset. It contains images of 632 persons, and each person has two images captured by two non-overlapping cameras with different viewpoints and illumination conditions.  The dataset is randomly split into two subsets of 316 persons each, for training and test respectively. And the procedure is repeated 10 times to get an average performance.

\subsubsection{Market-1501}
 Market-1501~\cite{zheng2015scalable} is currently the largest public available re-identification dataset, containing 32,668 detected bounding boxes of 1,501 persons, with each of them captured by six cameras at most and two cameras at least. Similar to the CUHK03 dataset, it also employs DPM  detector~\cite{felzenszwalb2010object}. We use the provided fixed training and test set, containing 750 and 751 identities respectively, to conduct experiments. 
 
 \begin{table*}[htbp]
 	\centering
 	\caption{Rank1, Rank5, Rank10 and Rank20 recognition rate (in \%) of various  methods on CUHK03 dataset with labeled setting. }
 	\begin{tabular}{ccccc}
 		\toprule
 		\textbf{Model} & \textbf{Rank1} & \textbf{Rank5} & \textbf{Rank10} & \textbf{Rank20} \\
 		\midrule
 		non-end-to-end CAN using Conv5 & 39.2 & 68.6 & 86.8 & 89.2 \\
 		non-end-to-end CAN using Max5 & 46.3 & 72.1 & 92.3 & 95.1 \\
 		\midrule
 		end-to-end	Avg pooled LSTM using Conv5 & 55.1 & 86.2 & 91.1 & 94.5 \\
 		end-to-end Max pooled LSTM using Conv5 & 54.4 & 85.1 & 91.7 & 93.8 \\
 		end-to-end Avg pooled LSTM using Max5 & 58.3 & 89.2 & 94.1 & 96.6 \\
 		end-to-end	Max pooled LSTM using Max5 & 57.9 & 88.9 & 93.3 & 95.2 \\
 		end-to-end	FC using Max5 & 53.8 & 86.5 & 90.1 & 93.5\\
 		end-to-end CAN using Conv5 & 63.8 & 91.0 & 95.2 & 97.1 \\
 		\midrule
 		\textbf{end-to-end CAN using Max5} & \textbf{72.3} & \textbf{93.8} & \textbf{98.4} & \textbf{99.2} \\
 		\bottomrule
 	\end{tabular}%
 	
 	\label{tab:conv_max}%
 \end{table*}%

\subsection{Implementation Details}\label{para_set}

\noindent \textbf{Pre-training of CNN:} As introduced in Sec. \ref{net_arch},  our proposed CAN includes CNN and attention components to learn comparative attention features. Specifically, CNN is exploited to learn global discriminative features  passed to attention components later. In this paper, we employ two classic CNNs: Alexnet\cite{krizhevsky2012imagenet} and VGG-16\cite{simonyan2014very}. And the difference of performance of our CAN by using different CNNs is validated in Sec.\ref{ana_mod}.  Before end-to-end training our CAN, we pre-train the CNN part. In details, we use the training set of each person re-identification dataset except for VIPeR to fine-tune the standard softmax classification CNN trained on ImageNet~\cite{deng2009imagenet} dataset. After pre-training, we remove the fully-connected classification layers and use the pre-trained CNN to initialize the CNN of our CAN model. Then the end-to-end training is performed on the specific training set of each person re-identification dataset. For the VIPeR dataset, considering the size of the training set is small, we use the training set of Market-1501 to pre-train the CNN and then use the training set of VIPeR to end-to-end train our CAN.
}

\noindent \textbf{Parameter Setting:\quad}We implement our network using Caffe~\cite{jia2014caffe} deep learning framework. The training of the CAN converges in roughly 8-10 hours on NVIDIA GeForce GTX TITAN X GPU. And it takes roughly 5-10 minutes for one split testing. In all of experiments, the dimensionality of the LSTM hidden state, the cell state, and the hidden layer  are set to 512 for CUHK01, CUHK03, VIPeR and Market-1501. The dimensionality of two layers within the MLP model ($f_{\text{init},c}$ and $f_{\text{init},h}$) in Eqn.~\eqref{c_ini} and \eqref{h_ini} are also set as 512 to initialize the LSTM memory states and hidden states. When the AlexNet is adopted, the images in all the datasets are resized to 227 $\times$ 227 while the images are resized to 224 $\times$ 224 when the VGG-16 is adopted to train our model. 
At the stage of pre-training of CNN, we perform stochastic gradient descent~\cite{bottou2012stochastic} to update the
weights. We start with a base learning rate of
$\eta^{(0)} = 0.01$ and gradually decrease it along with the training process
using an inverse policy: $\eta^{(k)} = \eta^{(0)}(1 + \gamma \cdot k)^{-p}$
where $\gamma = 10^{-4}$, $p = 0.75$, and $k$ is  index of the current mini-batch
iteration. We use a momentum of $\mu = 0.9$ and weight decay $\lambda = 5\times10^{-4}$. After the CNN feature learning network is pre-trained, we use the pre-trained model to initialize our end-to-end Comparative Attention Network (CAN). Here, we use the weight update parameter settings similar to those in the pre-training stage except that the initial learning rate is set to $\eta^{(0)} = 0.001$. As mentioned above, we adopt the on-line triplet selection method. Determined by cross-validation on CUHK03 dataset, the batchsize is set to 134 when CAN uses AlexNet and it is set to 66 when the VGG-16 is adopted. We chose the value of the margin parameter as $\alpha=0.3$ by cross-validation on CUHK03 dataset with labeled setting. In the experiments on other datasets, we also adopt the above parameter settings. As mentioned in  Sec.~\ref{lstm_can}, hidden states of LSTM at different time steps are concatenated as the final features passed to the normalization layer. Thus, we use 8 time steps and the extracted hidden states of the 2$^{nd}$, 4$^{th}$ and 8$^{th}$ time step in all experiments. It is illustrated in Fig.~\ref{fig:res_fig} (e) and Fig.~\ref{fig:difnum}, and is validated in Sec.~\ref{ana_mod}.   

\subsection{Data Augmentation}\label{data_aug}
In the training set, there exist much more negative pairs than positive pairs, which can lead to data imbalance and overfitting. To overcome this issue, we artificially augment the data by performing random 2D translation, similar to the processing in \cite{li2014deepreid}. For an original image of size $w \times h$, we sample
ten same-sized images around the image center, with translation drawn from a uniform distribution in the range [$-0.05w, 0.05w] \times [-0.05h, 0.05h]$. For all the datasets, we horizontally flip each image. In addition, because we use the on-line triplet selection method (see \ref{trip_sel}), we randomly shuffle the dataset in terms of their labels. Through this shuffle strategy, more triplets can be produced in a mini-batch. Specifically, we perform this operation ten rounds for each dataset.

\subsection{Analysis of the Proposed Model}\label{ana_mod}
\subsubsection{Ablation Study}
 In Sec.~III, we introduce our model architecture using CNN to learn global discriminative features.  To investigate the feature learned from which layer is more effective for our CAN model, we  use the CAN with AlexNet to conduct several experiments on the CUHK03 labeled dataset. We compare the performance of our model by using features learned from two different layers: $Conv5$ and $Max5$, which represent features from the 5$^{th}$ convolutional layer and from the 5$^{th}$ max pooling layer of AlexNet, respectively. Note, if $Conv5$ is used as a feature, the shape of the feature cube is 13 $\times$ 13 $\times$ 256, while if $Max5$ is used as a feature, the shape of the feature cube is 6 $\times$ 6 $\times$ 256. The experimental results on CUHK03 dataset are shown in Table~\ref{tab:conv_max}. From it, we observe that using $Max5$ can achieve better performance than using $Conv5$ as features. This may be because the $Max5$ can represent more abstract information for each person image, and provide more effective information for the subsequent comparative attention components. 

Moreover, different from \cite{sharma2015action},  our attention-based model is end-to-end trainable, which means that the comparative attention location maps can be obtained directly from the raw person images. So in Table \ref{tab:conv_max},  we also compare the performance of end-to-end CAN with the non-end-to-end CAN using pre-extracted CNN features. In the non-end-to-end CAN, the CNN features are extracted from the $Conv5$ and $Max5$ layers of pre-trained standard softmax classification CNN by using the pre-training method described in Sec.~\ref{para_set}. It shows that end-to-end CAN (``end-to-end CAN using Conv5'', ``end-to-end CAN using Max5'') can achieve better results than the one using pre-extracted CNN features (``non-end-to-end CAN using Conv5'', ``non-end-to-end CAN using Max5''). This is because the global feature learning and comparative attention components of the end-to-end version both participate in the process of comparing person images and updating the parameters. That is to say, the training loss would back-propagate not only the attention components but also the CNN part of our CAN. Otherwise, if the features are off-line pre-extracted and sent to the attention model, the CNN features may not contain enough comparative information since the CNN model is pre-trained using the network for the classification task. 

To further demonstrate the effectiveness of the comparative attention components in CAN, we also compare the performance of the proposed CAN with that of a similar architecture with masked input $\mathbf{A}_t$ of each LSTM replaced by the simple average pooling (``end-to-end Avg pooled LSTM using Conv5'', ``end-to-end Avg pooled LSTM using Max5'' ) or max pooling (``end-to-end Max pooled LSTM using Conv5'', ``end-to-end Max pooled LSTM using Max5'') over the CNN feature $\mathbf{X}_t$ in Fig.~\ref{fig:att_cp}. In other words, we use the same architecture illustrated in Fig.~\ref{fig:att_cp} except that none of the attention prediction mechanisms is contained in the model, and thus there is no softmax location map (attention map) $\mathbf{l}_t$ produced and all locations in a feature map have the same weight. Note that the LSTM model used here is also in an end-to-end form. From the results given in Table~\ref{tab:conv_max}, it is obvious that comparing positive pair and negative pair of each  person triplet and staying focusing on those more discriminative parts or locations can perform better than using the complete feature cube to discern different persons. Then we replace the LSTM part with two fully-connected layers with 512 dimensionality. From the results (``end-to-end FC using Max5''), we can observe that learning features in a recurrent way can indeed achieve better performance for person re-identification.

\begin{figure}[htb]
	\centering
	\includegraphics[width=7cm]{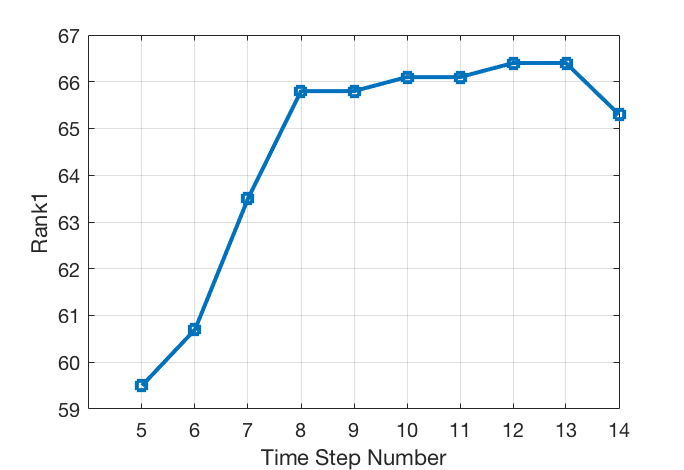}
	\caption{The Rank1 recognition rate (in \%) of our proposed CAN using different time step numbers, varying from 5 to 14, on CUHK03 dataset with labeled setting. 
	}
	
	\label{fig:difnum}
\end{figure}

 Additionally, we perform experiments with different time step numbers varying from 5 to 14. The performance is evaluated using the rank1 recognition rate. Here, we use the hidden states of all the time steps. The results are shown in the Fig.~\ref{fig:difnum}. We observe that the performance is gradually improved when the time step number increases from 5 to 8. However, further increasing step number from 8 to 14 does not bring significant improvement but increases the computational cost. So we choose the 8 time steps as it gives the best trade-off between performance and computational cost.

In the end, we also conduct a series of experiments to evaluate which time steps are chosen to be concatenated can achieve the best performance for our model. We use the following three settings: i) all the time steps (all 8 steps); ii) last time step (the 8$^{th}$ step); iii) the 2$^{nd}$, 4$^{th}$ and 8$^{th}$ time steps. The experiment shown in Fig.~\ref{fig:res_fig} (e) illustrates concatenating step-2/4/8 within our proposed CAN model gives best performance, rather than using all time steps. This is because discriminative information offered by the hidden states of adjacent time step may have redundancy. They are not very distinguishable from adjacent ones. Thus, combing all the hidden states may lead to over-smoothed features and lose discriminative information. In contrast, selecting features from the time step using our discovered interval can avoid over-smoothing and keep the necessary discriminative information at early steps. Moreover, using all the time steps can also cause the large dimensionality of the feature output by the concatenation layer (Eqn.~\eqref{concat}), which increases the computation cost. Under the second setting, our proposed CAN also can not achieve good performance because only using the last time step can not provide sufficient discriminative information and there is no integration of multiple local region features included.

\subsubsection{Effect of Different CNNs and Losses}
 
\begin{figure*}[th]
	\begin{center}
		\begin{tabular}{ccc}
			{\hspace{-3pt}}
			
			\includegraphics[height=0.26\linewidth,width=0.3\linewidth]{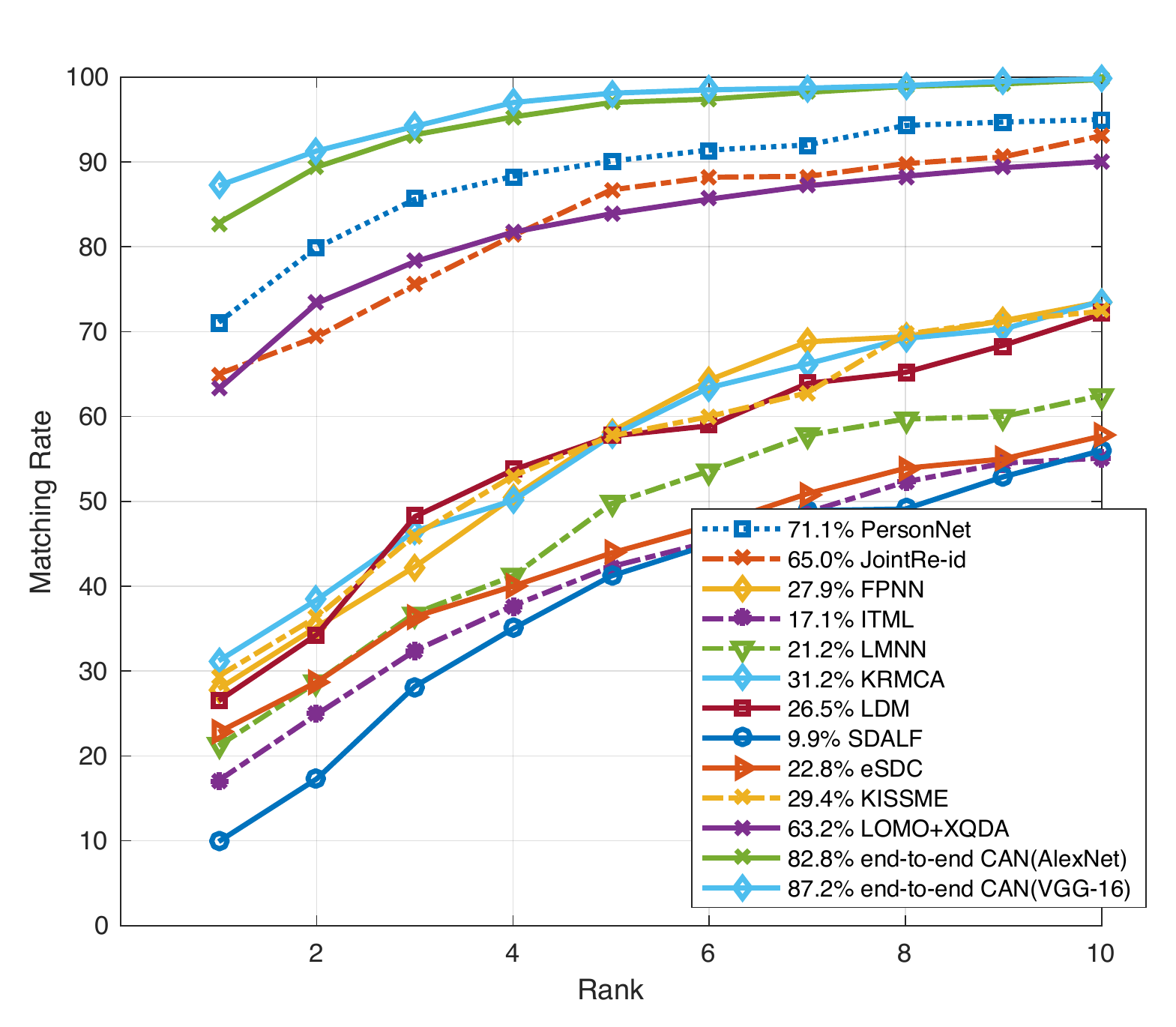}&		
			\includegraphics[height=0.25\linewidth,width=0.3\linewidth]{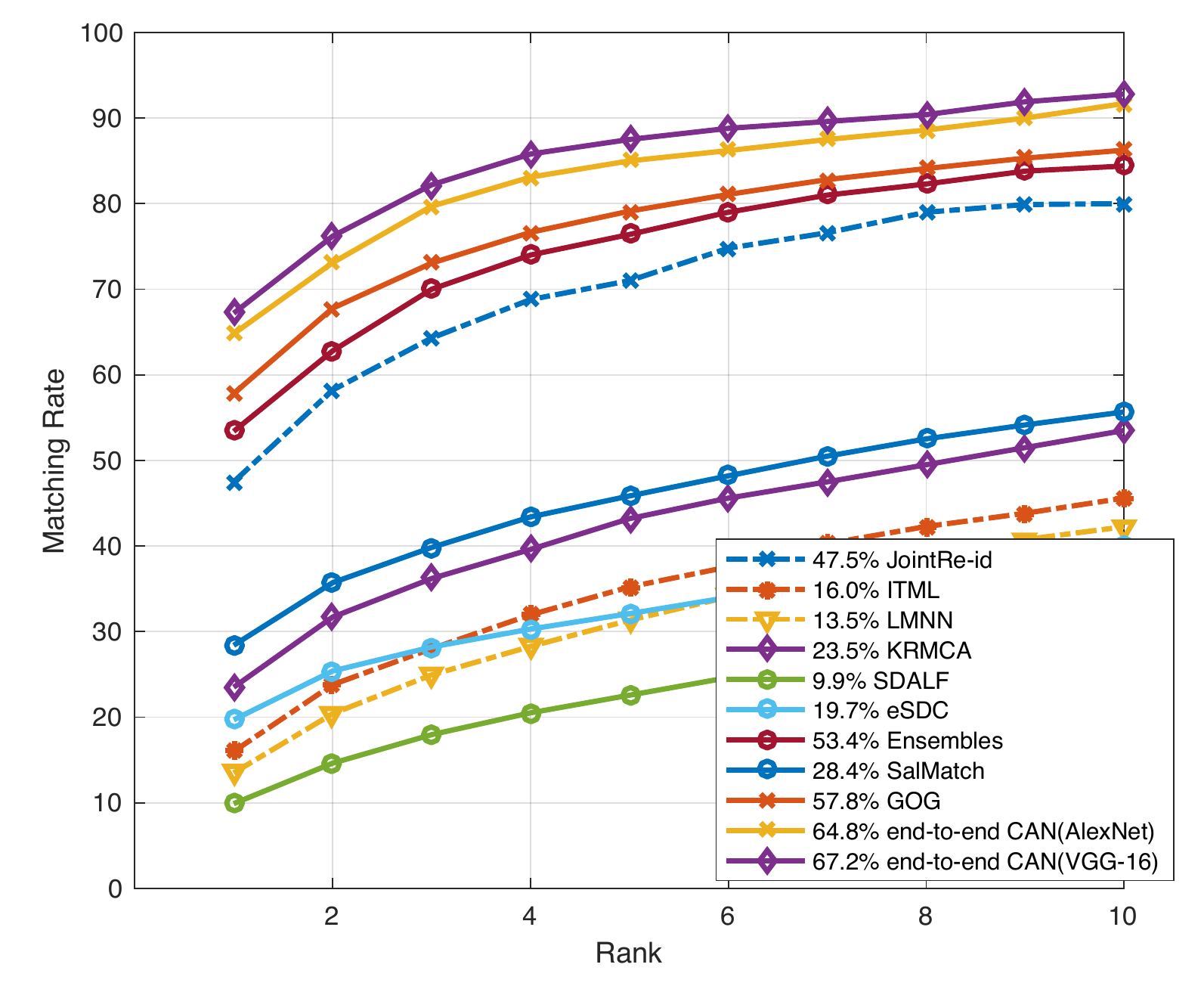}&	
			\includegraphics[height=0.255\linewidth, width=0.3\linewidth]{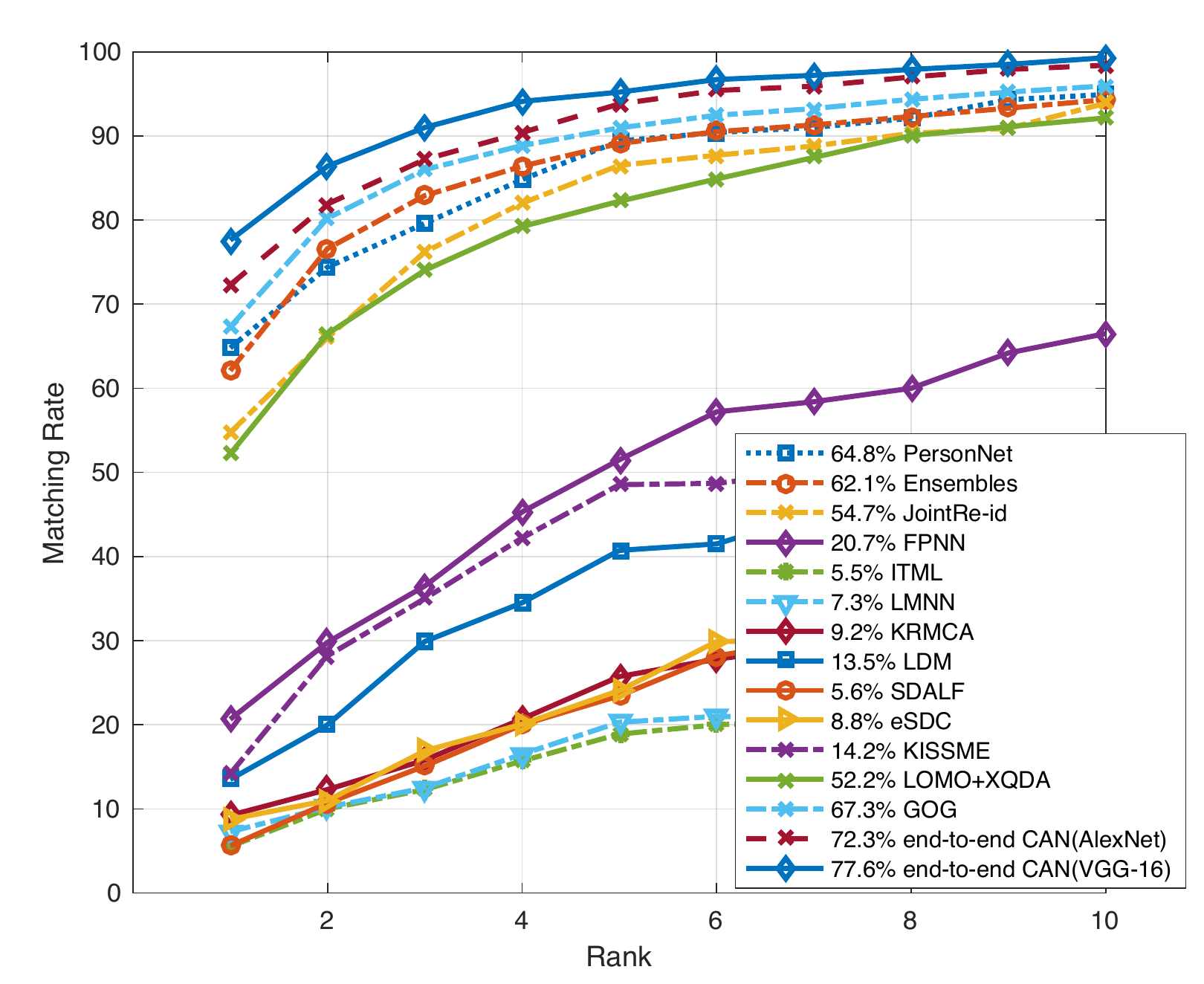}
			
			\\
			{\footnotesize{(a) CUHK01 (test=100) dataset}} & {\footnotesize{(b) CUHK01 (test=486) dataset}}& {\footnotesize{(c) CUHK03 labeled dataset}}
			\\
			\includegraphics[height=0.26\linewidth, width=0.3\linewidth]{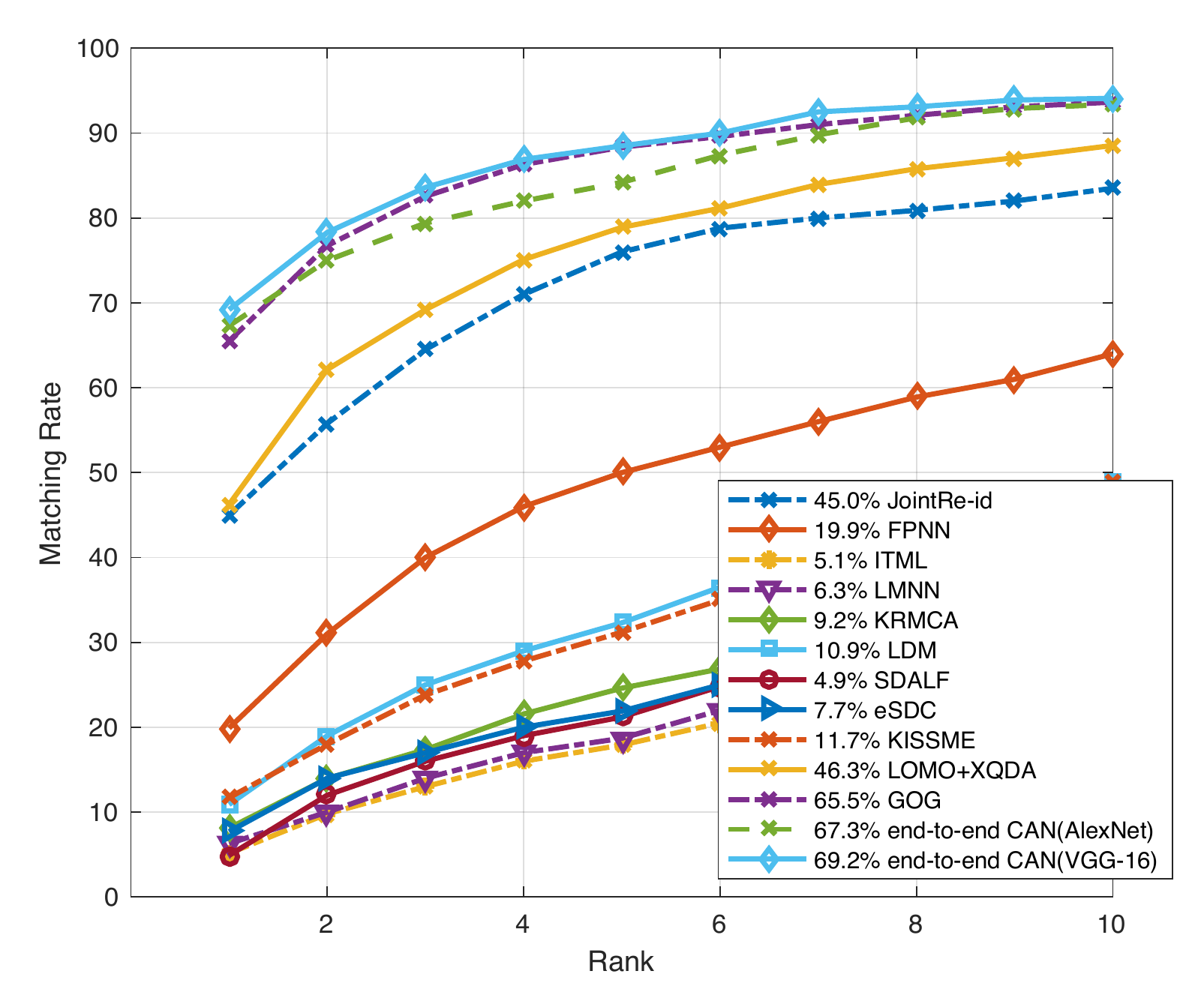}&
			\includegraphics[width=0.3\linewidth]{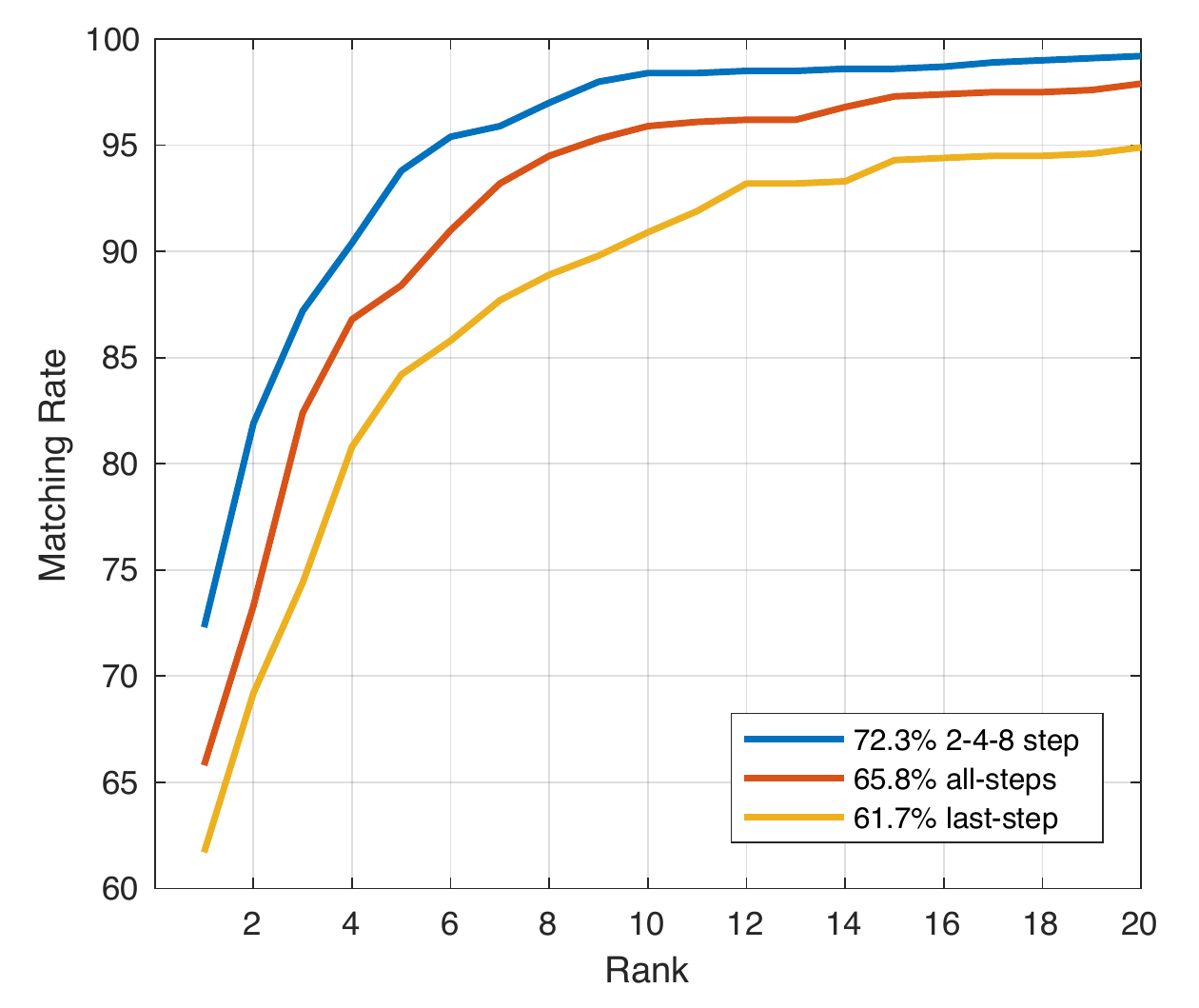}&		
			\includegraphics[height=0.26\linewidth, width=0.3\linewidth]{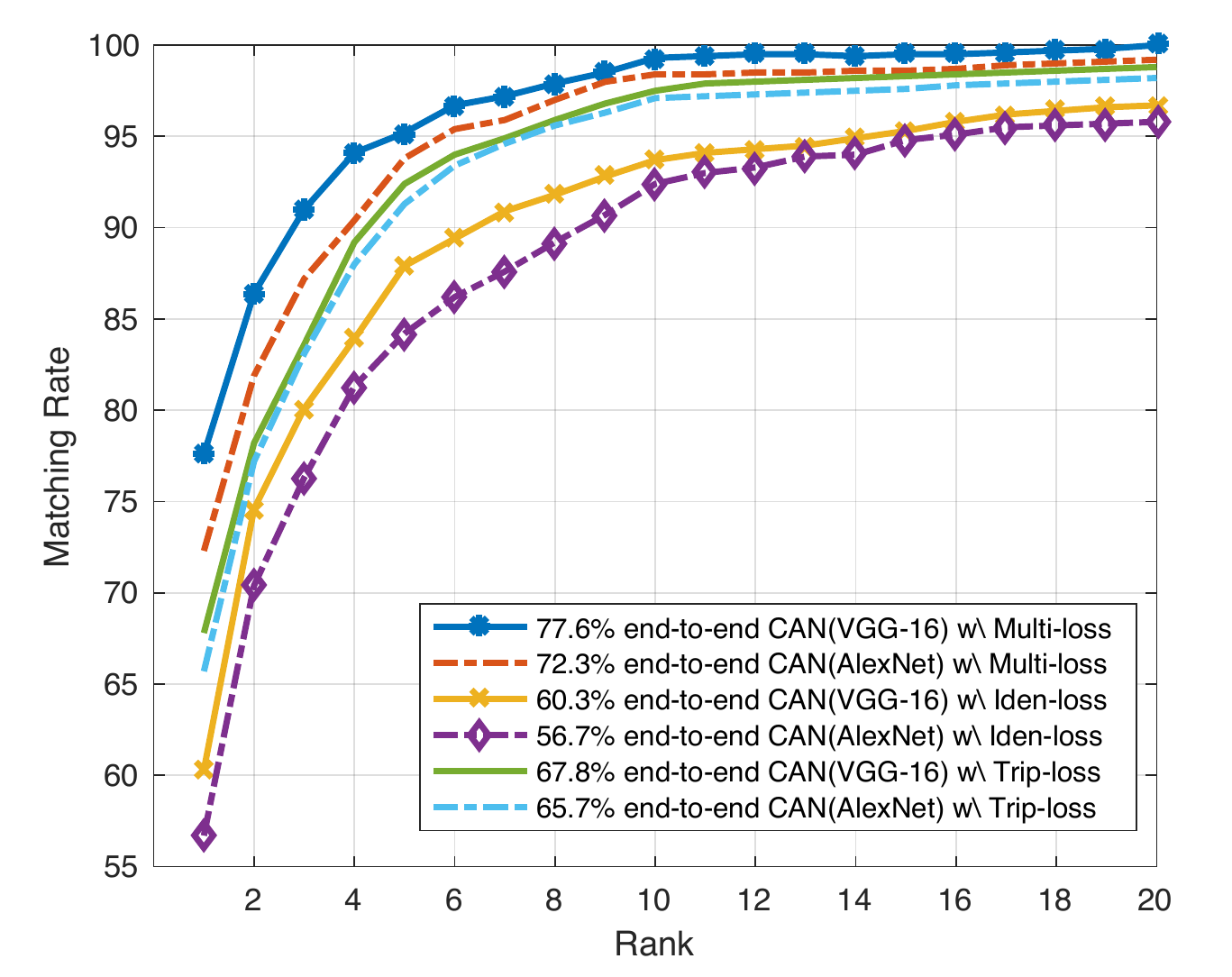}
			\\
			{\footnotesize{(d) CUHK03 detected dataset}} & {\footnotesize{(e) CUHK03 labeled dataset}} &{\footnotesize{(f) CUHK03 labeled dataset}}

		\end{tabular}
	\end{center}
	\caption{Performance comparison with state-of-the-art approaches using CMC curves on CHK01, CUHK03 labeled and CUHK03 detected datasets. (a) and (b) show comparisons of our method with previous methods on CUHK01 with 100 test IDs and 486 IDs, respectively. (c) and (d) show comparisons of our method with previous methods on CUHK03 labeled and detected, respectively. (e) gives comparison of our method with our own variations of concatenated time steps on CUHK03 labeled. (f) compares the performance of our method using different CNNs (AlexNet and VGG-16) and different losses (Multi-task loss, Triplet loss and Identity loss). Rank-1 matching rates are shown in the legend next to the method name. Our method beats the state-of-the-arts by a large margin, and gets the best performance when the 2$^{nd}$, 4$^{th}$ and 8$^{th}$ concatenated time steps are exploited. Our method can achieve the best performance by using VGG-16 as CNN and  multi-task loss as training loss. See \ref{ana_mod} and~\ref{cmp_state} for details. \textbf{Best viewed in 2 zoomed-in color pdf file.}
	}
	\label{fig:res_fig}
	\vspace{-1em}
\end{figure*}

In this subsection, we first compare the  performance of CANs using two different CNN models (\textit{i.e.}, AlexNet and VGG-16 respectively). When the VGG-16 is used, the feature learned from the 5$^{th}$ max pooling layer is passed to the attention components of  CAN. Therefore, the dimension of the passed feature cube is 7 $\times$ 7 $\times$ 512.  From the curves shown in Fig. \ref{fig:res_fig} (f), one can observe that the proposed CAN  can achieve better performance if it uses VGG-16 instead of AlexNet. This is because the VGG-16 model is a deeper network and is able to learn more discriminative  representations for person re-id. In Fig. \ref{fig:res_fig} (f), for both the CANs using AlexNet and VGG-16, one can also observe that using the multi-task loss can perform better  than using either triplet loss or identification loss individually. This implies that both the identification and the ranking information (conveyed in the triplet loss) are important for learning discriminative features through comparative attention. More importantly, the most effective features come from combining the power of the two loss functions.   

\subsection{Comparison with State-of-the-Art Methods}\label{cmp_state}
We compare our model  with the following state-of-the-art methods:  SDALF~\cite{farenzena2010person}, LMNN~\cite{hirzer2012person}, ITML~\cite{davis2007information}, KRMCA~\cite{liu2015kernelized}, LDM~\cite{guillaumin2009you} , eSDC~\cite{zhao2013unsupervised}, Metric Ensembles (Ensembles)~\cite{paisitkriangkrai2015learning}, KISSME~\cite{koestinger2012large}, JointRe-id~\cite{ahmed2015improved}, FPNN~\cite{li2014deepreid},  PersonNet~\cite{wu2016personnet}, LOMO+XQDA\cite{liao2015person}, FT-JSTL+DGD\cite{xiao2016learning}, TCP\cite{cheng2016person}, Embedding DM\cite{shi2016embedding}, LSSCDL\cite{Zhang_2016_CVPR} GOG\cite{matsukawa2016hierarchical}, SalMatch\cite{zhao2013person}, multi-HG\cite{an2016person}, DNS\cite{zhang2016learning}, 	TMA\cite{martinel2016temporal}, ROCCA\cite{LeAn:2015bn}, LMLF\cite{zhao2014learning}, CS\cite{shen2015person}, SCSP\cite{Chen_2016_CVPR}, RDs\cite{an2016personRD}, MLAPG\cite{liao2015efficient}, DSVR\_FSA\cite{tan2016dense}.

\begin{table}[htbp]
	\centering
	\caption{Comparison of our end-to-end CAN method's performance on the CUHK01 dataset with 100 test IDs to the state-of-the-art models.}
	\begin{tabular}{ccccc}
		\toprule
		\textbf{Method} & \textbf{Rank1} & \textbf{Rank5} & \textbf{Rank10} & \textbf{Rank20} \\
		\midrule
		JointRe-id\cite{ahmed2015improved} & 65.0    & 88.7  & 93.1 & 97.2 \\
		FPNN\cite{li2014deepreid}  & 27.9 & 58.2  & 73.5 & 86.3 \\
		ITML\cite{davis2007information}  & 17.1  & 42.3 & 55.1 & 71.7 \\
		LMNN\cite{hirzer2012person} & 21.2 & 49.7 & 62.5 & 78.6 \\
		KRMCA\cite{liu2015kernelized} & 31.2 & 57.7 & 73.6 & 86.1 \\
		LDM\cite{guillaumin2009you}   & 26.5 & 57.7 & 72.1 & 84.7 \\
		SDALF\cite{farenzena2010person} & 9.9   & 41.2 & 56.0  & 66.4 \\
		eSDC\cite{zhao2013unsupervised}  & 22.8 & 43.9 & 57.7 & 69.8 \\
		KISSME\cite{koestinger2012large} & 29.4  & 57.7 & 72.4 & 86.1 \\
		LOMO+XQDA\cite{liao2015person} &63.2&83.9 &90.1&94.2\\
		PersonNet\cite{wu2016personnet} & 71.1 & 90.1 & 95.0    & 98.1 \\
		Embedding DM\cite{shi2016embedding} & 86.6 &-&-&-\\
		\midrule
		{end-to-end CAN (AlexNet)} & {82.8} & {97.0} & {99.6} & {100.0} \\
		\textbf{end-to-end CAN (VGG-16)} & \textbf{87.2} & \textbf{98.2} &\textbf{99.8} & \textbf{100.0} \\
		\bottomrule
	\end{tabular}%
	\label{tab:CUHK01}%
\end{table}%

\subsubsection{Results on CUHK01 and CUHK03}\label{res_cuhk0103}

These two datasets consist of  thousands of training samples. Table~\ref{tab:CUHK01} and Fig.~\ref{fig:res_fig} (a) show results on CUHK01 with 100 test IDs. Our method beats all compared methods at low ranks. For the CUHK01 with 486 test IDs, our method can also beat other compared methods. Although our attention-based CAN is also a kind of saliency method, it beats the other saliency-based methods\cite{zhao2013person, zhao2013unsupervised, zhao2014learning} by a large margin. As for CUHK03, there are two
settings: manually cropped person images and person images produced by DPM detector. Obviously, the performance on the latter one appears lower than that on the former, as shown in Fig.~\ref{fig:res_fig} (c), Fig.~\ref{fig:res_fig} (d), Table~\ref{tab:CUHK03_lab} and Table~\ref{tab:CUHK03_detec}. However, the images produced by the detector can
also  reflect the algorithms in the real world. It can be
seen from Fig.~\ref{fig:res_fig} (c) and Table~\ref{tab:CUHK03_lab} that, as expected, on this large
dataset, some other deep learning based methods, such as FT-JSTL+DGD\cite{xiao2016learning}, can achieve similar performance with  millions of parameters or become much more competitive. However, with
the detector boxes, our method is less affected, especially for the Rank1, and outperforms other approaches including deep learning based ones by a large margin. We suppose that the performance is not affected too much, possibly  because our model could accurately attend to different discriminative parts of images and integrate their information which is robust to the influence brought by the detector.

\begin{table}[htbp]
	\centering
	\caption{Comparison of our end-to-end CAN method's performance on the CUHK01 dataset with 486 test IDs to the state-of-the-art models.}
	\begin{tabular}{ccccc}
		\toprule
		\textbf{Method} & \textbf{Rank1} & \textbf{Rank5} & \textbf{Rank10} & \textbf{Rank20} \\
		\midrule
		FT-JSTL+DGD\cite{xiao2016learning} & 66.6 &- &- &-\\
		SDALF\cite{farenzena2010person} & 9.9   & 22.6 & 30.3  & 41.0 \\
		ITML\cite{davis2007information}  &16.0 & 35.2& 45.4 & 59.8  \\
		LMNN\cite{hirzer2012person} &13.5 & 31.3 & 42.3 & 54.1\\
		KRMCA\cite{liu2015kernelized} & 23.5 & 43.2 & 53.5 & 63.2 \\
		eSDC\cite{zhao2013unsupervised} &19.7&32.7 &40.3& 50.6\\
		LSSCDL\cite{Zhang_2016_CVPR} &66.0 &- &- &-\\
		DSVR\_FSA\cite{tan2016dense} &33.5 &50.9 &61.0 &71.0\\
		MidLevel\cite{zhao2014learning}&34.3 &55.1 &65.0 &75.0\\
		TCP\cite{cheng2016person} &53.7 & 84.3&91.0 &96.3\\
		Ensembles\cite{paisitkriangkrai2015learning} &53.4&76.4&84.4&90.5\\
		JointRe-id\cite{ahmed2015improved} &47.5&71.0&80.0&-\\
		SalMatch\cite{zhao2013person} & 28.4 &45.8&55.7&67.9\\
		ROCCA\cite{LeAn:2015bn} &29.8&-&67.8&77.0\\
		GOG\cite{matsukawa2016hierarchical} &57.8 &79.1 &86.2 &92.1\\
		multi-HG\cite{an2016person} &64.4&-&90.6&94.6\\
		DNS\cite{zhang2016learning}&\textbf{69.1}&86.9&91.8&95.4\\
		\midrule
		{end-to-end CAN (AlexNet)} & {64.8} & {84.7} & {91.7} & {96.8} \\
		\textbf{end-to-end CAN (VGG-16)} & {67.2} & \textbf{87.3} & \textbf{92.5} & \textbf{97.2} \\
		\bottomrule
	\end{tabular}%
	\label{tab:CUHK01_486}%
\end{table}%

\begin{table}[htbp]
	\centering
	\caption{Comparison of our end-to-end CAN method's performance on the CUHK03 dataset with labeled setting to the state-of-the-art models.}
	\begin{tabular}{ccccc}
		\toprule
		\textbf{Method} & \textbf{Rank1} & \textbf{Rank5} & \textbf{Rank10} & \textbf{Rank20} \\
		\midrule
		Ensembles\cite{paisitkriangkrai2015learning} & 62.1  & 89.1 & 94.3  & 97.8 \\
		JointRe-id\cite{ahmed2015improved} & 54.7 & 86.4 & 91.5  & 97.3 \\
	    FPNN\cite{li2014deepreid}  & 20.7 & 51.5 & 68.7 & 83.1 \\
		ITML\cite{davis2007information}  & 5.5  & 18.9& 30.0 & 44.2\\
		LMNN\cite{hirzer2012person}  & 7.3 & 21.0    & 32.0 & 48.9 \\
		KRMCA\cite{liu2015kernelized} & 9.2  & 25.7 & 35.1 & 53.0 \\
		LDM\cite{guillaumin2009you}   & 13.5 & 40.7 & 52.1 & 70.8 \\
		SDALF\cite{farenzena2010person} & 5.6   & 23.5 & 36.1 & 52.0 \\
		eSDC\cite{zhao2013unsupervised}  & 8.8  & 24.1 & 38.3 & 53.4 \\
		KISSME\cite{koestinger2012large} & 14.2 & 48.5 & 52.6 & 70.0 \\
		LOMO+XQDA\cite{liao2015person} &52.2 &82.2 &92.1 &96.3\\
		PersonNet\cite{wu2016personnet} & 64.8  & 89.4  & 94.9 & 98.2 \\
		GOG\cite{matsukawa2016hierarchical}&67.3&91.1&96.0&98.8\\
		Embedding DM\cite{shi2016embedding} & 61.3 &-&-&-\\
		DNS\cite{zhang2016learning} &62.6&90.1&94.8&98.1\\
		FT-JSTL+DGD\cite{xiao2016learning} &75.3 & - & -  &-\\
		\midrule
		{end-to-end CAN(AlexNet)} & {72.3} & {93.8} & 98.4 & 99.2 \\
		\textbf{end-to-end CAN(VGG-16)} & \textbf{77.6} & \textbf{95.2} & \textbf{99.3} & \textbf{100.0 }\\
		\bottomrule
	\end{tabular}%
	\label{tab:CUHK03_lab}%
\end{table}%

\begin{table}[htbp]
	\centering
	\caption{Comparison of our end-to-end CAN method's performance on the CUHK03 dataset with detected setting to the state-of-the-art models.}
	\begin{tabular}{ccccc}
		\toprule
		\textbf{Method} & \textbf{Rank1} & \textbf{Rank5} & \textbf{Rank10} & \textbf{Rank20} \\
		\midrule
		JointRe-id\cite{ahmed2015improved} & 45.0 & 76.0 & 83.5 & 93.2 \\
		FPNN\cite{li2014deepreid}  & 19.9 & 50.0 & 64.0 & 78.5 \\
		ITML\cite{davis2007information}  & 5.1  & 17.9 & 28.3 & 43.1 \\
		LMNN\cite{hirzer2012person}  & 6.3  & 18.7 & 29.1 & 45.0 \\
		KRMCA\cite{liu2015kernelized} & 8.1  &20.3  &33.0  &50.0  \\
		LDM\cite{guillaumin2009you}   & 10.9 &32.3 &48.8  &65.6  \\
		SDALF\cite{farenzena2010person} & 4.9  &21.2 & 35.1  &48.4  \\
		eSDC\cite{zhao2013unsupervised}  & 7.7  &21.9  &35.0  &50.1  \\
		KISSME\cite{koestinger2012large} & 11.7  &31.2 &49.0  & 65.6 \\
		LOMO+XQDA\cite{liao2015person}&46.3 &78.9 &88.6 &94.3\\
		GOG\cite{matsukawa2016hierarchical}&65.5&88.4&93.7&97.6\\
		Embedding DM\cite{shi2016embedding} & 52.1 &-&-&-\\
		DNS\cite{zhang2016learning} &54.7 &86.8 &94.8 &95.2\\
		\midrule
			{end-to-end CAN(AlexNet)} & {67.3} & {84.2} & 93.4 & 95.2 \\
		\textbf{end-to-end CAN(VGG-16)} & \textbf{69.2} & \textbf{88.5} & \textbf{94.1} & \textbf{97.8 }\\
		\bottomrule
	\end{tabular}
	\label{tab:CUHK03_detec}%
\end{table}%

\subsubsection{Results on Market-1501}\label{res_market}

Market-1501 is a large and realistic dataset since it was captured in a scene of crowded
supermarket with complex environment. Besides, it contains several natural detector errors as the person images were collected by applying the automatic DPM detector. Each person is captured by  by six cameras at most in Market-1501~\textemdash~the number of cameras is significantly larger than the CUHK01 and CUHK03 datasets. Therefore, the relationships between person pairs are more complicated. We compare the performance of our proposed CAN model against state-of-the-art results under both the single query and multi-query settings and with both evaluation metrics in Table~\ref{tab:market}. The performance of baseline methods given in \cite{zheng2015scalable} is not very competitive, probably because BoW (Bag-of-Word) features the authors used are not robust enough. From Table \ref{tab:market}, we can observe that the rank1 performance of our CAN model is slightly lower  than the  current best method (DNS\cite{zhang2016learning}) under single query setting, but it still achieves the mAP as high as 35.9\%. Under the multiple query setting, our method can provide the new state-of-the-art. This shows that our model performs comparable to other methods in more complex multi-camera person re-identification tasks, benefiting from the inherent attention and comparison mechanism.

\begin{table}[htbp]
	\centering
	\caption{Comparison of our end-to-end CAN method’s performance on the Market-1501 dataset with both s ery and multiple query setting to the state-of-the-art models.}
	\begin{tabular}{c|cc|cc}
		\toprule
		\textbf{Query} &\multicolumn{2}{c|}{\textbf{SingleQ}}&\multicolumn{2}{c}{\textbf{MultipleQ}}\\
		\midrule
		\textbf{Method} & \textbf{Rank1} & \textbf{mAP} & \textbf{Rank1} & \textbf{mAP}\\
		\midrule
		SDALF\cite{farenzena2010person}  & 20.5 & 8.2 &29.2 &13.8\\
		eSDC\cite{zhao2013unsupervised}  & 33.5 & 13.5 &42.5 &18.4 \\
		LOMO+TMA\cite{martinel2016temporal}& 47.9 & 22.3&-&- \\
		Zheng et al.\cite{zheng2015scalable} & 34.4  & 14.1 &42.6 &19.5\\
		PersonNet\cite{wu2016personnet} & 37.2 & 18.6 &-&-\\
			SCSP\cite{Chen_2016_CVPR} &51.9&26.4&-&-\\
			DNS\cite{zhang2016learning}&\textbf{61.1} &35.7 &71.6 &46.0\\
			
		\midrule
		{end-to-end CAN(AlexNet)} & {55.1} & {30.3} &65.4&42.2\\
		\textbf{end-to-end CAN(VGG-16)} & {60.3} & \textbf{35.9} &\textbf{72.1}&\textbf{47.9}\\
		\bottomrule
	\end{tabular}%
	\label{tab:market}%
\end{table}%

\subsubsection{Results on VIPeR}
Compared with the aforementioned three datasets, VIPeR\cite{gray2007evaluating} is one of the most challenging, since it has 632 people but with various poses, viewpoints, image resolutions, and lighting conditions. And each person has only one image under each camera. From Table \ref{tab:viper}, one can see that our proposed method (``end-to-end CAN(VGG-16)") can beat  most of the compared state-of-the-arts except SCSP\cite{Chen_2016_CVPR} exploiting hand-crafted features. This is because the size of the training set of VIPeR is so small that our deep learning-based method is easy to overfit to the training set with millions of parameters. To overcome the overfitting problem, we perform the data augmentation introduced in Sec. \ref{data_aug} and fine-tune our proposed CAN based on the model pre-trained on other large person re-identification datasets such as Market-1501. However, the data are still insufficient to generate enough  person triplets to train our CAN model.  Compared with the experimental results on large datasets, such as CUHK03 and Market-1501, discussed in Sec.\ref{res_cuhk0103} and Sec. \ref{res_market}, the methods using low-level hand-crafted  features are easier to achieve comparable results with deep learning-based methods on the small dataset. Therefore, we also conduct another experiment combining the output features of CAN and LOMO (LOcal Maximal Occurrence representation)\cite{liao2015person} which is a kind of classic low-level feature containing both color and texture features. The combination is performed by simply concatenating the CAN feature vector and the LOMO feature vector.  The effectiveness is verified by the results (``CAN(VGG-16)+LOMO'') shown in Table \ref{tab:viper}.
\begin{table}[htbp]
	\centering
	\caption{Comparison of our end-to-end CAN method's performance on the VIPeR dataset to the state-of-the-art models.}
	\begin{tabular}{ccccc}
		\toprule
		\textbf{Method} & \textbf{Rank1} & \textbf{Rank5} & \textbf{Rank10} & \textbf{Rank20} \\
		\midrule
	ROCCA\cite{LeAn:2015bn}&30.4&-&75.6&86.6\\
	FT-JSTL+DGD\cite{xiao2016learning} &38.6&-&-&-\\
	JointRe-id\cite{ahmed2015improved} &34.8&64.5&78.5&89.1\\
	PCCA\cite {mignon2012pcca} &19.3&48.9&64.9&80.3\\
	SDALF\cite{farenzena2010person}  &19.9  &38.4 &49.4&66.0 \\
	eSDC\cite{zhao2013unsupervised}  &26.3 &46.4 &58.6&72.8 \\
	SalMatch]\cite{zhao2013person} &30.2&52.3&66.0&73.4\\
	KRMCA\cite{liu2015kernelized}&23.2&54.8 &72.2&85.5\\
	KISSME\cite{koestinger2012large}&19.6&48.0&62.2&77.0\\
	MidLevel+LADF\cite{zhao2014learning}&43.4&73.0&84.9&93.7\\
	Ensembles\cite{paisitkriangkrai2015learning}&45.9&-&-&-\\
	LMLF\cite{zhao2014learning}&29.1&52.3&66.0&79.9\\
	CS\cite{shen2015person}&34.8&68.7&82.3&91.8\\
	DSVR\_FSA\cite{tan2016dense}&29.4 &50.7 &62.0 &75.0\\
	TCP\cite{cheng2016person}&47.8&74.7&84.8&91.1\\
	SCSP\cite{Chen_2016_CVPR} &53.5&82.6&91.5&\textbf{96.6}\\
	multi-HG\cite{an2016person}&44.7&-&83.0&92.4\\
	RDs\cite{an2016personRD} &33.3&-&78.4&88.5\\
	TMA\cite{martinel2016temporal} &39.9&-&81.3&91.5\\
	LOMO+XQDA\cite{liao2015person}&40.00&68.9&80.5&91.1\\
	GOG\cite{matsukawa2016hierarchical}&49.7&79.7&88.7&94.5\\
	MLAPG\cite{liao2015efficient}&40.7&-&82.3&92.4\\
	DNS\cite{zhang2016learning}&51.2&82.1&90.5&95.9\\
	
		\midrule
		{end-to-end CAN(AlexNet)} &41.5 &72.6&83.2&92.7 \\
		{end-to-end CAN(VGG-16)} &47.2&79.2&89.2&95.8 \\
		\textbf{ CAN(VGG-16)+LOMO} &\textbf{54.1}&\textbf{83.1}&\textbf{91.8}&96.4 \\
		\bottomrule
	\end{tabular}
	\label{tab:viper}%
\end{table}%

\subsection{Visualization of Attention Maps and Discussions}\label{vis}
\begin{figure*}[ht]
	\begin{center}
		\begin{tabular}{ccc}
			{\hspace{-3pt}}
			
			\includegraphics[width=0.4\linewidth]{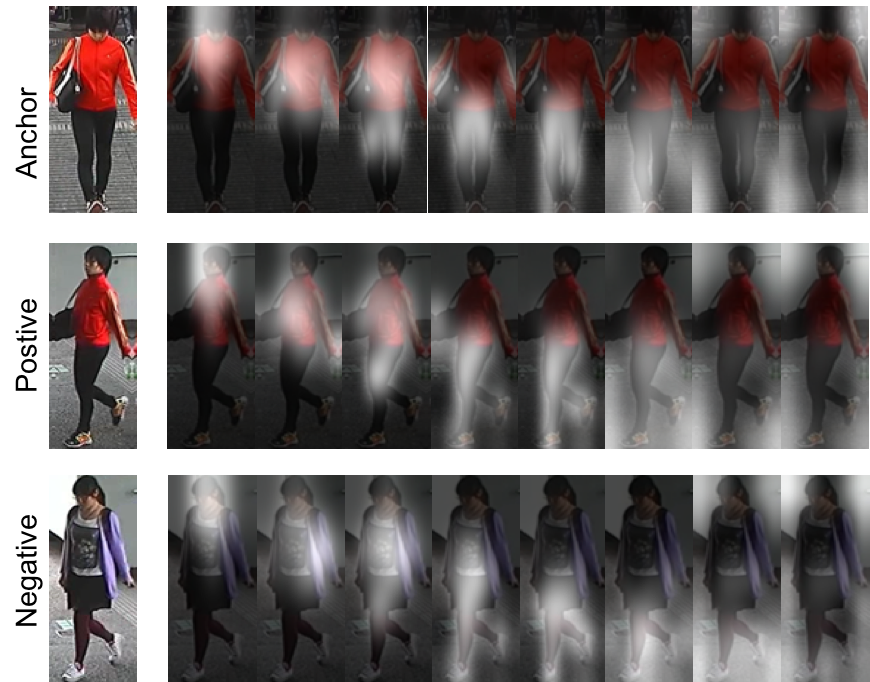}&		
			\includegraphics[width=0.4\linewidth]{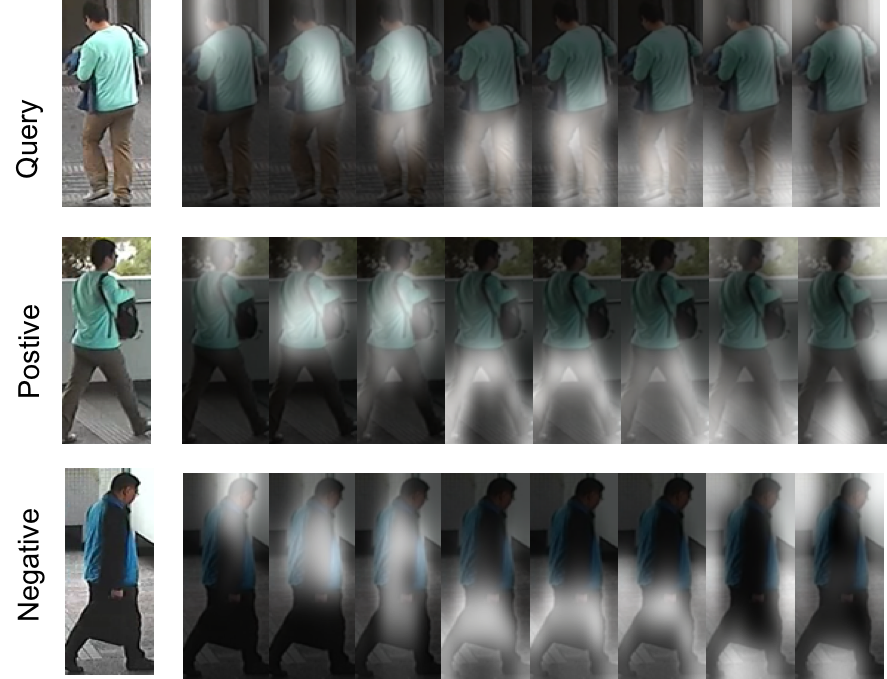}&	
			\\
			{\footnotesize{(a) Training samples from CUHK01 dataset}}  & 
			{\footnotesize{(b) Testing samples from CUHK01 dataset}}&
			\\
			\includegraphics[width=0.4\linewidth]{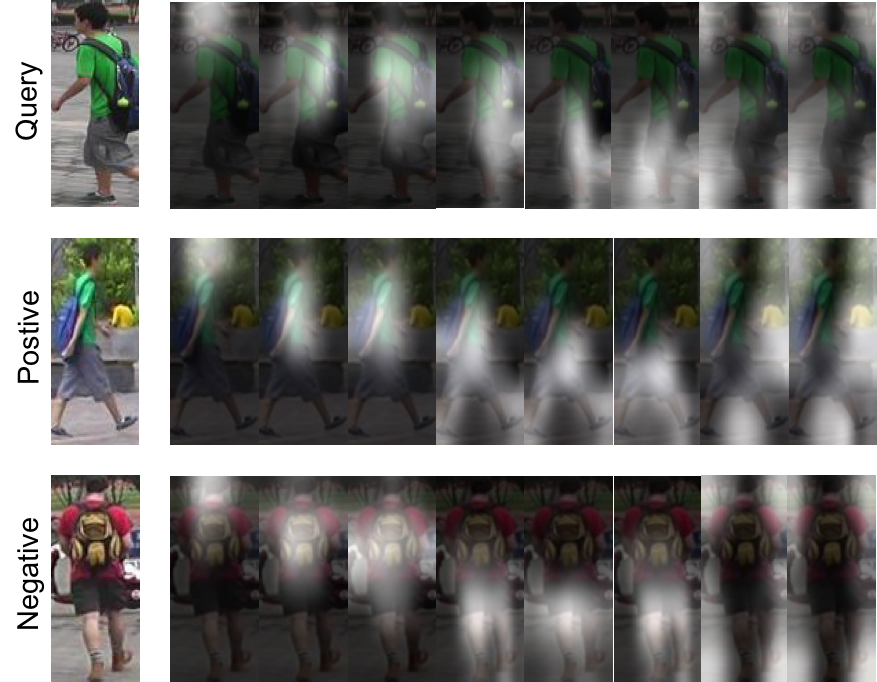}&
			\includegraphics[width=0.4\linewidth]{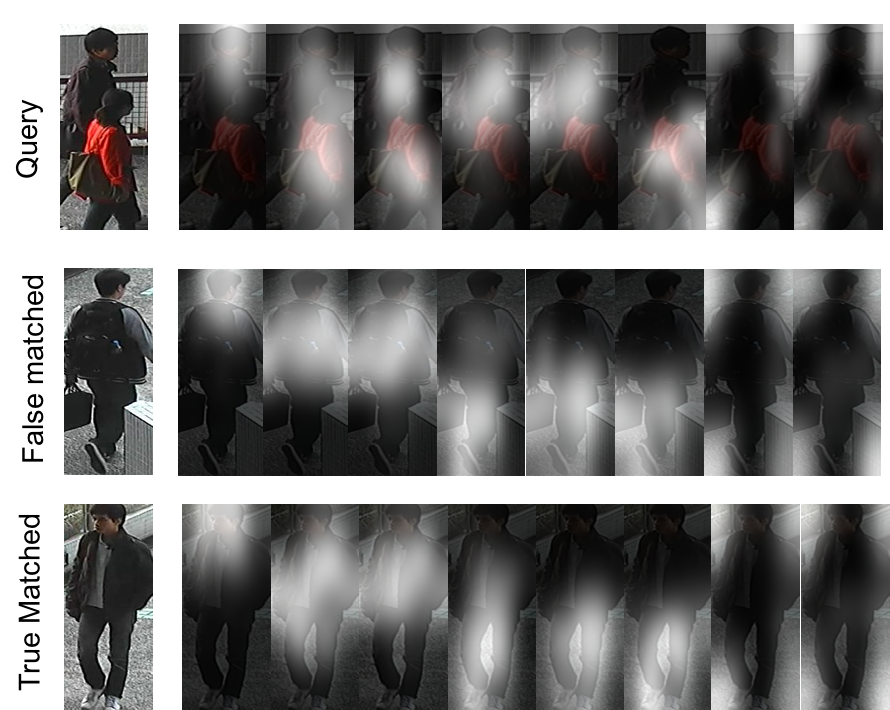}&
			\\
			 {\footnotesize{(c)  Testing samples from Market-1501 dataset}}& {\footnotesize{(d) Failure  samples from CUHK03 labeled dataset}}&

		\end{tabular}
	\end{center}
	\caption{ Attention maps learned by our network for different training and testing person samples in CUHK01 (test=100) dataset and CUHK03 labeled dataset. (a) and (b)  show the triplets of training and testing samples from CUHK01 (test=100) dataset, respectively. (c) shows the testing samples from Market-1501 dataset under single query setting. (d) shows some failure cases of our model for CUHK03 labeled dataset.  In Fig.~\ref{fig:att_exp} (a), the triplet of training samples is randomly selected from one batch. And in  Fig.~\ref{fig:att_exp} (b) and (c),  the positive sample is ranked at top 1 in re-identification results while the negative sample is randomly selected from those ranked at bottom 10 in re-identification results.  For both  of triplet samples from training and testing set, the comparative attention is often focused on the discriminative parts based on which the comparison can tell the persons are the same (for positive pairs) and are different (for negative pairs). }
	\label{fig:att_exp}
	\vspace{-1em}
\end{figure*}

In Fig.~\ref{fig:att_exp}, we visualize some comparative attention maps produced by our network for both training samples (Fig.~\ref{fig:att_exp} (a)) and  testing samples (Fig.~\ref{fig:att_exp} (b)) from CUHK01 dataset with 100 test IDs. In Fig.~\ref{fig:att_exp} (a), the triplet of training samples is randomly selected from one batch. And in  Fig.~\ref{fig:att_exp} (b),  the positive sample is ranked at top 1 in re-identification results while the negative sample is randomly selected from those ranked at bottom 10 in re-identification results. 

 In the attention maps generated on training samples (In Fig.~\ref{fig:att_exp} (a)), we can see that the model is able to focus on different parts of the person images at different time steps. The attention often starts from the head parts in the triplet sample images, and then is focused on the upper parts of the body at the second step. In the next several time steps, the attention gradually shifts on the lower parts of the images in a triplet. That is to say, our CAN model often focuses on the discriminative parts based on which the comparison can tell the persons are the same (for positive pairs) and are different (for negative pairs). Thus, the attention maps learned by our CAN model only represent which parts within the triplet are used by our CAN for comparison. Take the Fig.~\ref{fig:att_exp} (a) in which the query person wearing the red cloth for example. At the second step, the attention of CAN is focused on the upper body part. Thus intuitively our CAN compares what kinds of clothes the persons wearing to make decision: the positive pair of persons wear red cloth and the negative pairs of persons wear purple cloth. This part attended at this time step can tell whether the persons are the same or different. But the cloth information is not very reliable. Therefore, CAN proceeds to collect information with different attention in the following steps. At a single time step, if the attention does not focus on the corresponding local regions for both matched pair and unmatched pair, then the comparison is meaningless. For example, comparing the head part of one person with the leg part of other person cannot tell whether they are the same person or not. Note that, the model does not always  attend to the foreground.  We can see that some background is attended to at last two steps, which means the background can also provide information to assist matching persons correctly.  The similar attention map changes of testing samples can also be observed in Fig.~\ref{fig:att_exp} (b).  

The comparative attention maps of testing samples from Market-1501 dataset under single query setting are also visualized in Fig.~\ref{fig:att_exp} (c). Different from the CUHK01 dataset in which the person images are manually cropped, the person images are automatically detected by the DPM detector in the  Market-1501 dataset. Therefore, more difficult cases, such as misalignment and body part missing, are contained due to the detector errors. Through the Fig.~\ref{fig:att_exp} (c), we can see that the attention maps of testing samples from Market-1501 also change in the similar way with that of CUHK01 dataset.     

In Fig.~\ref{fig:att_exp} (d), we also visualize failure cases of our proposed CAN model in terms of the generated attention maps based on the comparison mechanism. We can observe that our model fails to focus on the same parts of positive person image pairs. This is partially because there is more than one person in the  query image and this person is occluded heavily by other persons. This extremely hard scenario poses a big challenge to our CAN model as it can not exactly compare person image pairs and decide which local region should be selected. This phenomenon can also be observed from the generated attention maps of the query sample, false matched sample and true matched sample in Fig.~\ref{fig:att_exp} (d).

\section{Conclusion}
In this work, we introduced a novel visual attention model that is formulated as a triplet recurrent neural network which takes several glimpses of triplet images of persons and dynamically generates comparative attention location maps for person re-identification. We conducted extensive experiments on three public available person re-identification datasets to validate our method. Experimental results demonstrated that our model outperforms other state-of-the-art methods in most cases, and verified that our comparative attention model is beneficial for
the recognition accuracy in person matching.


%

\ifCLASSOPTIONcaptionsoff
  \newpage
\fi



%
\bibliography{reid}{}
\bibliographystyle{IEEEtran}
\end{document}